\documentclass[10pt,twocolumn,letterpaper]{article}

\usepackage{cvpr}
\usepackage{times}
\usepackage{epsfig}
\usepackage{graphicx}
\usepackage{amsmath}
\usepackage{amssymb}
\usepackage{mathtools}

\usepackage{tabu}
\usepackage{empheq}

\usepackage{placeins}
\usepackage{afterpage}

\usepackage[format=plain,labelformat=simple,labelsep=period,font=small,skip=4pt,compatibility=false]{caption}
\usepackage[font=scriptsize,skip=2pt]{subcaption}

\usepackage{tabularx}
\usepackage{setspace}
\usepackage{nicefrac}
\usepackage{booktabs}
\usepackage[super]{nth}

\usepackage{siunitx}
\usepackage[pagebackref=true,breaklinks=true,letterpaper=true,colorlinks,bookmarks=false]{hyperref}
\cvprfinalcopy 

\def\cvprPaperID{2485} 

\ifcvprfinal\fi

\newcommand{\qed}{\nobreak \ifvmode \relax \else
      \ifdim\lastskip<1.5em \hskip-\lastskip
      \hskip1.5em plus0em minus0.5em \fi \nobreak
      \vrule height0.75em width0.5em depth0.25em\fi}

\begin{document}

\title{Self-Supervised Feature Learning by Learning to Spot Artifacts}

\author{Simon Jenni \qquad Paolo Favaro\\
University of Bern, Switzerland\\
{\tt\small \{jenni,favaro\}@inf.unibe.ch}}


\maketitle

\begin{abstract}
We introduce a novel self-supervised learning method based on adversarial training. Our objective is to train a discriminator network to distinguish real images from images with synthetic artifacts, and then to extract features from its intermediate layers that can be transferred to other data domains and tasks. 
To generate images with artifacts, we pre-train a high-capacity autoencoder and then we use a \emph{damage and repair} strategy: First, we freeze the autoencoder and \emph{damage} the output of the encoder by randomly dropping its entries. Second, we augment the decoder with a \emph{repair} network, and train it in an adversarial manner against the discriminator. The repair network helps generate more realistic images by inpainting the dropped feature entries. To make the discriminator focus on the artifacts, we also make it predict what entries in the feature were dropped. 
We demonstrate experimentally that features learned by creating and spotting artifacts achieve state of the art performance in several benchmarks.
\end{abstract}

\section{Introduction}
Recent developments in deep learning have demonstrated impressive capabilities in learning useful features from images \cite{krizhevsky2012imagenet}, which could then be transferred to several other tasks \cite{girshickICCV15fastrcnn,girshick14CVPR,renNIPS15fasterrcnn,sharif2014cnn}.
These systems rely on large annotated datasets, which require expensive and time-consuming human labor.
To address these issues \emph{self-supervised learning} methods have been proposed \cite{doersch2015unsupervised,noroozi2016unsupervised,pathak2016context,wang2015unsupervised,zhang2016colorful}.
These methods learn features from images without annotated data. Some introduce a \emph{pretext task} through the following strategy: one withholds some information about the input data and then trains a network to recover it. For example, some methods withhold image regions \cite{pathak2016context}, color \cite{zhang2016colorful} or both grayscale and color  values \cite{zhang2016split}; others have widthheld the location of patches \cite{doersch2015unsupervised,noroozi2016unsupervised}, or additional external information such as egomotion \cite{jayaraman2015learning}.
In self-supervised learning the main challenge is to define a pretext task that relates the most to the final applications of the learned features.
\begin{figure}[t]
\begin{center}
\setlength{\fboxsep}{0pt}
\setlength{\fboxrule}{1pt}
\fcolorbox{red}{white}{\includegraphics[width=0.19\linewidth,trim={0cm 0cm 0cm 0cm},clip]{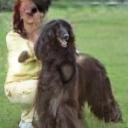}}%
\fcolorbox{red}{white}{\includegraphics[width=0.19\linewidth,trim={0cm 0cm 0cm 0cm},clip]{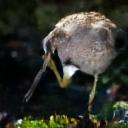}}%
\fcolorbox{red}{white}{\includegraphics[width=0.19\linewidth,trim={0cm 0cm 0cm 0cm},clip]{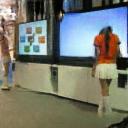}}%
\fcolorbox{green}{white}{\includegraphics[width=0.19\linewidth,trim={0cm 0cm 0cm 0cm},clip]{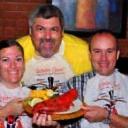}}%
\fcolorbox{green}{white}{\includegraphics[width=0.19\linewidth,trim={0cm 0cm 0cm 0cm},clip]{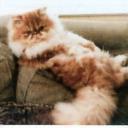}}\\\vspace{-.4mm}%
\fcolorbox{red}{white}{\includegraphics[width=0.19\linewidth,trim={0cm 0cm 0cm 0cm},clip]{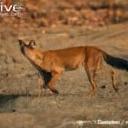}}%
\fcolorbox{red}{white}{\includegraphics[width=0.19\linewidth,trim={0cm 0cm 0cm 0cm},clip]{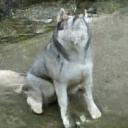}}%
\fcolorbox{red}{white}{\includegraphics[width=0.19\linewidth,trim={0cm 0cm 0cm 0cm},clip]{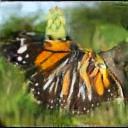}}%
\fcolorbox{green}{white}{\includegraphics[width=0.19\linewidth,trim={0cm 0cm 0cm 0cm},clip]{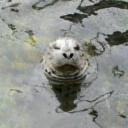}}%
\fcolorbox{green}{white}{\includegraphics[width=0.19\linewidth,trim={0cm 0cm 0cm 0cm},clip]{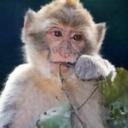}}\\\vspace{-.4mm}%
\fcolorbox{red}{white}{\includegraphics[width=0.19\linewidth,trim={0cm 0cm 0cm 0cm},clip]{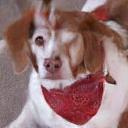}}%
\fcolorbox{red}{white}{\includegraphics[width=0.19\linewidth,trim={0cm 0cm 0cm 0cm},clip]{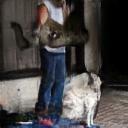}}%
\fcolorbox{green}{white}{\includegraphics[width=0.19\linewidth,trim={0cm 0cm 0cm 0cm},clip]{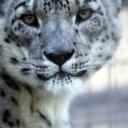}}%
\fcolorbox{green}{white}{\includegraphics[width=0.19\linewidth,trim={0cm 0cm 0cm 0cm},clip]{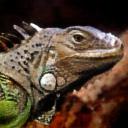}}%
\fcolorbox{green}{white}{\includegraphics[width=0.19\linewidth,trim={0cm 0cm 0cm 0cm},clip]{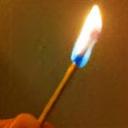}}%
\vspace{-2mm}
\end{center}
   \caption{A mixture of real images (green border) and images with synthetic artifacts (red border). Is a good object representation necessary to tell them apart?}
\label{fig:artifacts}
\end{figure}

Towards this goal, we propose to learn features by classifying images as real or with artifacts (see Figure~\ref{fig:artifacts}). We aim at creating image artifacts, such that a model capable of spotting them would require an accurate representation of objects and thus build features that could transfer well to tasks such as object classification, detection and segmentation. 
A first approach to create artifacts is to use inpainting algorithms \cite{bertalmio2000image,denton2016semi}. Besides being computationally inefficient on large inpainting regions, these methods are unsuitable for training because they may introduce low-level statistics that a neural network could easily learn to detect. This could limit what the network learns about objects.
Therefore, instead of editing images at the pixel level, we tamper with their feature representation and create corrupt images so that the texture artifacts are locally unnoticeable, but globally incorrect, as illustrated in Figure~\ref{fig:artifacts}.

To generate artifacts we first train an autoencoder to reproduce images. Then, we randomly drop entries from the encoded feature (at the bottleneck) so that some information about the input image is lost. We then add a \emph{repair} neural network to the decoder to help it render a realistic image. The repair network inpaints the feature representations at every layer of the decoder, but its limited capacity does not allow it to fully recover the missing information. In this way we obtain an image with artifacts that cannot be detected through local analysis. We then train a discriminator to distinguish real from corrupt images. Moreover, we also make the discriminator output a mask to indicate what feature entries were dropped. This implicitly helps the discriminator focus on the important details in the image. We also use the true mask to restrict the scope of the repair network to the features corresponding to the dropped entries. This limits the ability of the repair network to replace missing information in a globally consistent manner. 
The repair network and the discriminator are then trained in an adversarial fashion. However, in contrast to other adversarial schemes, notice that our repair network is designed not to completely confuse the discriminator. Finally, we transfer features from the discriminator, since this is the model that learns an approximation of the distribution of images.

Our contributions can be summarized as follows: 1) a novel feature learning framework based on detecting images with artifacts, which does not require human annotation; 2) a method to create images with non-trivial artifacts; 3) our features achieve state of the art performance on several transfer learning evaluations (ILSVRC2012 \cite{imagenet_cvpr09}, Pascal VOC \cite{everingham2010pascal} and STL-10 \cite{coates2011analysis}).





%
%
%
%
%
%
%
%
%
%

\section{Prior Work}

This work relates to several topics in machine learning: adversarial networks, autoencoders and self-supervised learning, which we review briefly here below.\\
\textbf{Adversarial Training.}
The use of adversarial networks has been popularized by the introduction of the generative adversarial network (GAN) model by Goodfellow \etal~\cite{goodfellow2014generative}.
Radford \etal~\cite{radford2015unsupervised} introduced a convolutional version of the GAN along with architectural and training guidelines in their DCGAN model. 
Since GANs are notoriously difficult to train much work has focused on heuristics and techniques to stabilize training \cite{salimans2016improved,radford2015unsupervised}.
Donahue \etal~\cite{donahue2016adversarial} extend the GAN model with an encoder network that learns the inverse mapping of the generator. Their encoder network learns features that are comparable to contemporary unsupervised methods. They show that transferring the discriminator features in the standard GAN setting leads to significantly worse performance. In contrast, our model demonstrates that by limiting the capabilities of the generator network to local alterations the discriminator network can learn better visual representations. \\
\textbf{Autoencoders.}
The autoencoder model is a common choice for unsupervised representation learning  \cite{hinton1994autoencoders}, with two notable extensions: the denoising autoencoder \cite{vincent2010stacked} and the variational autoencoder (VAE) \cite{kingma2013auto}.
Several combinations of autoencoders and adversarial networks have recently been introduced. Makhzani \etal~\cite{makhzani2015adversarial} introduce adversarial autoencoders,
which incorporate concepts of VAEs and GANs by training the latent hidden space of an autoencoder to match (via an adversarial loss) some pre-defined prior distribution. 
In our model the autoencoder network is separated from the generator (\ie, repair network) and we do not aim at learning semantically meaningful representations with the encoder. \\
\textbf{Self-supervised Learning.}
A recently introduced paradigm in unsupervised feature learning uses either naturally occurring or artificially introduced supervision as a means to learn visual representations.
The work of Pathak \etal~\cite{pathak2016context} combines autoencoders with an adversarial loss for the task of inpainting (\ie, to generate the content in an image region given its context) with state-of-the-art results in semantic inpainting. Similarly to our method, Denton \etal \cite{denton2016semi} use the pretext task of inpainting and transfer the GAN discriminator features for classification. However, we remove and inpaint information at a more abstract level (the internal representation), rather than at the raw data level. This makes the generator produce more realistic artifacts that are then more valuable for feature learning. Wang and Gupta~\cite{wang2015unsupervised} use videos and introduce supervision via visual tracking of image-patches. Pathak \etal \cite{pathakCVPR17learning} also make use of videos through motion-based segmentation. The resulting segmentation is then used as the supervisory signal. 
Doersch \etal~\cite{doersch2015unsupervised} use spatial context as a supervisory signal by training their model to predict the relative position of two random image patches.
Noroozi and Favaro~\cite{noroozi2016unsupervised} build on this idea by solving jigsaw puzzles.
Zhang \etal~\cite{zhang2016colorful} and Larsson \etal \cite{larsson2017colorproxy} demonstrate good transfer performance with models trained on the task of image colorization, \ie, predicting color channels from luminance. The split-brain autoencoder~\cite{zhang2016split} extends this idea by also predicting the inverse mapping.
The recent work of Noroozi \etal~\cite{noroozi2017representation} introduces counting of visual primitives as a pretext task for representation learning. Other sources of natural supervision include ambient sound \cite{owens2016ambient}, temporal ordering \cite{misra2016shuffle} and physical interaction \cite{pinto2016curious}. 
We compare our method to the above approaches and show that it learns features that, when transferred to other tasks, yield a higher performance on several benchmarks.


\section{Architecture Overview}

We briefly summarize the components of the architecture that will be introduced in the next sections.
Let $\mathbf{x}$ be a training image from a dataset and $\hat{\mathbf{x}}$ be its version with artifacts. 
As model, we use a neural network consisting of the following components (see also Figure~\ref{fig:architecture}):

\begin{figure*}[t]
\begin{center}
  \includegraphics[width=0.95\linewidth,trim={0.cm 0cm 0.cm 0cm},clip]{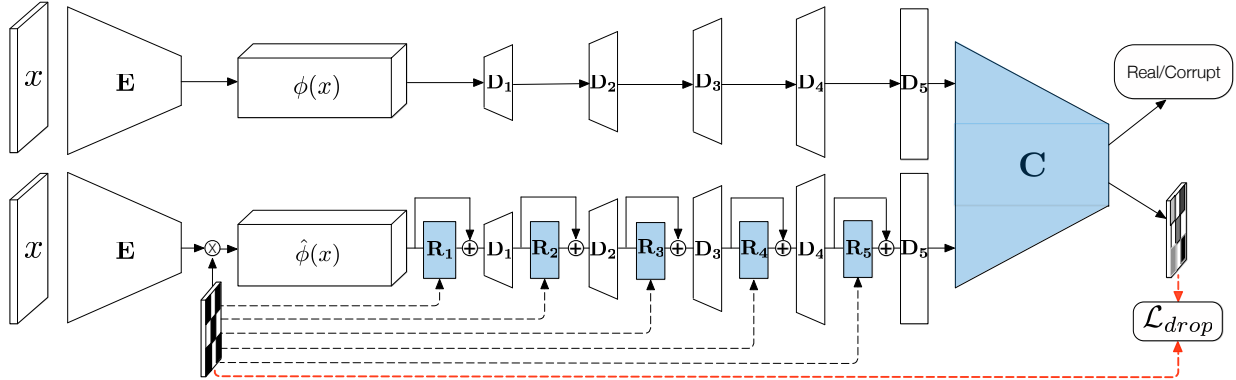}
\end{center}
   \caption{The proposed architecture. Two autoencoders $\{E,D_1,D_2,D_3,D_4,D_5\}$ output either real images (top row) or images with artifacts (bottom row). A discriminator $C$ is trained to distinguish them. The corrupted images are generated by masking the encoded feature $\phi(\mathbf{x})$ and then by using a repair network $\{R_1,R_2,R_3,R_4,R_5\}$ distributed across the layers of the decoder. The mask is also used by the repair network to change only the dropped entries of the feature (see Figure~\ref{fig:repairblock} for more details). The discriminator and the repair network (both shaded in blue) are trained in an adversarial fashion on the real/corrupt classification loss. The discriminator is also trained to output the mask used to drop feature entries, so that it learns to localize all artifacts.}
\label{fig:architecture}
\end{figure*}

\noindent\textbf{1.} Two \emph{autoencoder} networks $\{E,D_1,D_2,D_3,D_4,D_5\}$, where $E$ is the encoder and $D = \{D_1,D_2,D_3,D_4,D_5\}$ is the decoder, pre-trained to reproduce high-fidelity real images $\mathbf{x}$; $\phi(\mathbf{x})$ is the output of the encoder $E$ on $\mathbf{x}$;\\
\noindent\textbf{2.} A spatial mask $\Omega$ to be applied to the feature output of $E$; the resulting masked feature is denoted $\hat \phi(\mathbf{x}) = \Omega \odot \phi(\mathbf{x}) + (1-\Omega) \odot (u \ast  \phi(\mathbf{x}))$, where $u$ is some uniform spatial kernel so that $u \ast  \phi(\mathbf{x})$ is a feature average, and $\odot$ denotes the element-wise product in the spatial domain (the mask is replicated along the channels);\\ 
\noindent\textbf{3.} A \emph{discriminator} network $C$ to classify $\mathbf{x}$ as real images and $\hat{\mathbf{x}}$ as fake; we also train the discriminator to output the mask $\Omega$, so that it learns to localize all artifacts;\\
\noindent\textbf{4.} A \emph{repair} network $\{R_1,R_2,R_3,R_4,R_5\}$ added to the layers of one of the two decoder networks; the output of a layer $R_i$ is masked by $\Omega$ so that it affects only masked features.\\~\\
The repair network and the discriminator are trained in an adversarial fashion on the  real/corrupt classification loss.

\section{Learning to Spot Artifacts}

Our main objective is to train a classifying network (the discriminator) so that it learns an accurate distribution of real images. Prior work \cite{radford2015unsupervised} showed that a discriminator trained to distinguish real from fake images develops features with interesting abstraction capabilities. In our work we build on this observation and exploit a way to control the level of corruption of the fake images (see Sec.~\ref{sec:repair}).
Thus, we train a classifier to discriminate between real and corrupt images (see Sec.~\ref{sec:adv}). 
As illustrated earlier on, by solving this task we hope that the classifier learns features suitable for other tasks such as object classification, detection and segmentation. In the next sections we describe our model more in detail, and present the design choices aimed at avoiding learning trivial features.

\begin{figure}[t]
\begin{center}
\begin{subfigure}[h]{0.195\linewidth}
	\includegraphics[width=\linewidth,trim={0cm 0cm 0cm 0cm},clip]{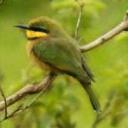}%
\hspace{-\linewidth}\includegraphics[width=0.25\linewidth,trim={0cm 0cm 0cm 0cm},clip]{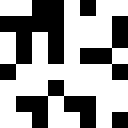}%
\end{subfigure}\hspace{.4mm}%
\begin{subfigure}[h]{0.195\linewidth}
	\includegraphics[width=\linewidth,trim={0cm 0cm 0cm 0cm},clip]{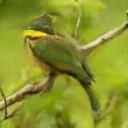}%
\end{subfigure}\hspace{.4mm}%
\begin{subfigure}[h]{0.195\linewidth}
	\includegraphics[width=\linewidth,trim={0cm 0cm 0cm 0cm},clip]{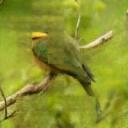}%
\end{subfigure}\hspace{.4mm}%
\begin{subfigure}[h]{0.195\linewidth}
	\includegraphics[width=\linewidth,trim={0cm 0cm 0cm 0cm},clip]{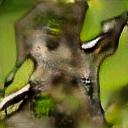}%
\end{subfigure}\hspace{.4mm}%
\begin{subfigure}[h]{0.195\linewidth}
	\includegraphics[width=\linewidth,trim={0cm 0cm 0cm 0cm},clip]{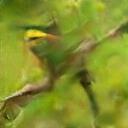}%
\end{subfigure}\\
\begin{subfigure}[h]{0.195\linewidth}
	\includegraphics[width=\linewidth,trim={0cm 0cm 0cm 0cm},clip]{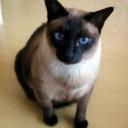}%
\hspace{-\linewidth}\includegraphics[width=0.25\linewidth,trim={0cm 0cm 0cm 0cm},clip]{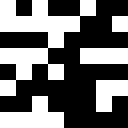}%
        \caption{}
\end{subfigure}\hspace{.4mm}%
\begin{subfigure}[h]{0.195\linewidth}
	\includegraphics[width=\linewidth,trim={0cm 0cm 0cm 0cm},clip]{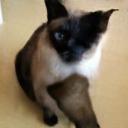}%
        \caption{}
\end{subfigure}\hspace{.4mm}%
\begin{subfigure}[h]{0.195\linewidth}
	\includegraphics[width=\linewidth,trim={0cm 0cm 0cm 0cm},clip]{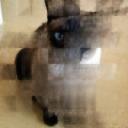}%
        \caption{}
\end{subfigure}\hspace{.4mm}%
\begin{subfigure}[h]{0.195\linewidth}
	\includegraphics[width=\linewidth,trim={0cm 0cm 0cm 0cm},clip]{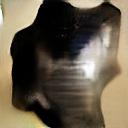}%
        \caption{}
\end{subfigure}\hspace{.4mm}%
\begin{subfigure}[h]{0.195\linewidth}
	\includegraphics[width=\linewidth,trim={0cm 0cm 0cm 0cm},clip]{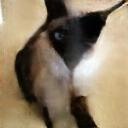}%
        \caption{}
\end{subfigure}
\end{center}
   \caption{Two examples of corrupt images obtained from our \emph{damage \& repair} network. (a) shows two original images from the ImageNet dataset. At the bottom-left corner of those images we show the masks applied to the encoded feature $\phi(\mathbf{x})$. These masks drop on average about $50\%$ of the encoded feature. (b) shows the output corrupt images. The repair network assists the decoder in inpainting texture that is only locally unnoticeable. However, at the global scale the objects are no longer recognizable as valid instances. (c) shows the output of the decoder when the repair network is not active. In this case the artifacts are very visible and easy to detect by exploiting low-level statistics. (d) shows the output of the decoder when the repair network is not masked. The repair network is then able to change the image more globally. This has a negative effect on the discriminator as it fails to predict the mask. (e) shows an example where the images are fed through the \emph{damage \& repair} network twice. This results in even more artifacts than in (b).}
\label{fig:repair}
\end{figure}

\subsection{The Damage \& Repair Network}
\label{sec:repair}
In our approach we would like to be able to create corrupt images that are not too unrealistic, otherwise, a classifier could distinguish them from real images by detecting only low-level statistics (\eg, unusual local texture patterns). At the same time, we would like to have as much variability as possible, so that a classifier can build a robust model of real images.

To address the latter concern we randomly corrupt real images of an existing dataset. To address the first concern instead of editing images at the pixel-level, we corrupt their feature representation and then partly repair the corruption by fixing only the low-level details.
We encode an image $\mathbf{x}$ in a feature $\phi(\mathbf{x})\in \mathbf{R}^{M\times N \times L}$, where $\phi(\mathbf{x}) = E(\mathbf{x})$, and then at each spatial coordinate in the $M\times N$ domain we randomly drop all the $L$ channels with a given probability $\theta\in(0,1)$. This defines a binary mask matrix $\Omega\in \{0,1\}^{M\times N}$ of what feature entries are dropped ($\Omega_{ij}=0$) and which ones are preserved ($\Omega_{ij}=1$). 
The dropped feature channels are replaced by the corresponding entries of an averaged feature computed by convolving $\phi(\mathbf{x})$ with a large uniform kernel $u$. As an alternative, we also replace the dropped feature channels with random noise. Experimentally, we find no significant difference in performance between these two methods.
The mask $\Omega$ is generated online during training and thus the same image $\mathbf{x}$ is subject to a different mask at every epoch.
If we fed the corrupt feature directly to the decoder $D$ the output would be extremely unrealistic (see Figure~\ref{fig:repair}~(c)). Thus, we introduce a \emph{repair} network that partially compensates for the loss of information due to the mask. The repair network introduces repair layers $R_i$ between layers of the decoder $D_i$. These layers receive as input the corrupt feature or the outputs of the decoder layers and are allowed to fix only entries that were dropped. More precisely, we define the input to the first decoder layer $D_1$ as 
\begin{align}
\hat \phi(\mathbf{x}) + (1-\Omega)\odot R_1(\hat \phi(\mathbf{x})),
\end{align}
where $\hat \phi(\mathbf{x}) = \Omega \odot \phi(\mathbf{x}) + (1-\Omega) \odot (u\ast \phi(\mathbf{x}))$ and $u$ is a large uniform filter.
At the later layers $D_2$, $D_3$, and $D_4$ we upsample the mask $\Omega$ with the nearest neighbor method and match the spatial dimension of the corresponding intermediate output, \ie, we provide the following input to each layer $D_i$ with $i=2,3,4$
\begin{align}
D_{i-1} + (1-U_{i-1}(\Omega))\odot R_i(D_{i-1}),
\end{align}
where $U_{i-1}$ denotes the nearest neighbor upsampling to the spatial dimensions of the output of $D_{i-1}$.
Finally, we also design our encoder $E$ so that it encodes features that are spatially localized and with limited overlap with one another. To do so, we define the encoder $E = \{E_1,E_2,E_3,E_4,E_5\}$ with five layers where $E_1$ uses $3\times 3$ convolutional filters with stride $1$ and the remaining four layers $E_2$, $E_3$, $E_4$, and $E_5$ use $2\times 2$ convolutional filters with stride $2$.
\begin{figure}[t]
\begin{center}
	\includegraphics[width=\linewidth,trim={0cm 0cm 0cm 0cm},clip]{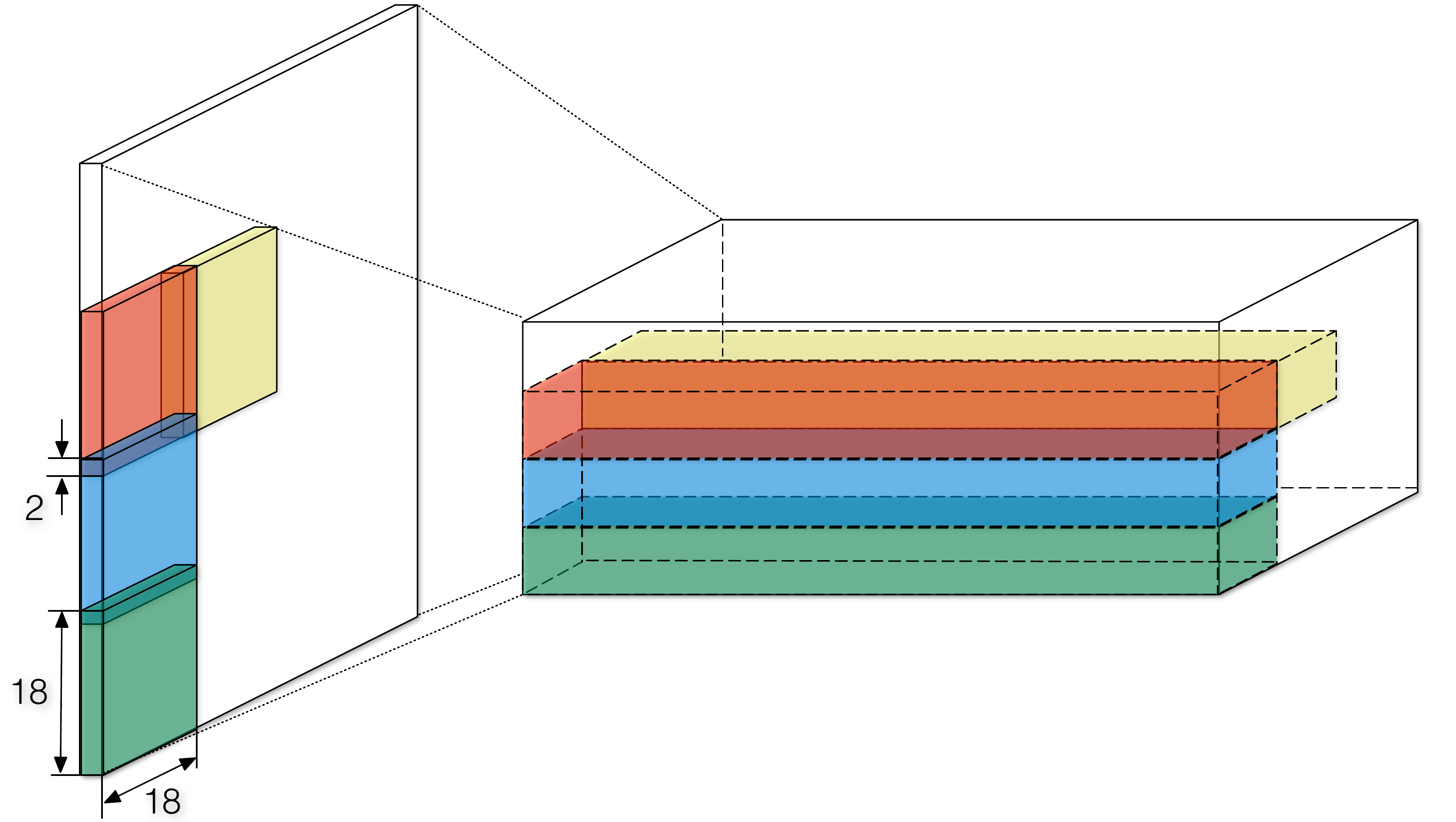}%
\end{center}
   \caption{Encoder design. On the left we show the input image and on the right we show the corresponding feature encoding $\phi(\mathbf{x})$. We use $5$ layers, where the first one uses $3\times 3$ convolutional filters with stride $1$ and the remaining $4$ use $2\times 2$ convolutional filters with stride $2$. As can be observed, this results in almost separate receptive fields for each feature entry of $\phi(\mathbf{x})$. Each entry corresponds to a $18\times 18$ pixels patch in the original input and overlaps $2$ pixels with the neighboring patches ($1$ pixel each side). This encoding ensures a strong spatial locality to each entry of the encoded features.}
\label{fig:encoder}
\end{figure}
As shown in Figure~\ref{fig:encoder}, this design limits the receptive field of each encoded feature entry to a $18\times 18$ pixels patch in the input image with a $2$ pixels overlap with neighboring patches. Thus, dropping one of these feature entries is essentially equivalent to dropping one of the $18\times 18$ pixels patches in the input image.

In Figure~\ref{fig:repair} we show two examples of real images and corresponding corrupt images obtained under different conditions. Figure~\ref{fig:repair}~(a) shows the original images and the mask $\Omega$, where $M,N=8$, as inset on the bottom-left corner. The dropping probability is $\theta=0.5$. Figure~\ref{fig:repair}~(b) shows the resulting corrupt images obtained by our complete architecture. Notice that locally it is impossible to determine whether the image is real or corrupt. Only by looking at a larger region, and by exploiting prior knowledge of what real objects look like, it is possible to determine that the image is corrupt. For comparison purposes we show in Figure~\ref{fig:repair}~(c) and (d) the corresponding corrupt images obtained by disabling the repair network and by using the repair network without the mask restriction, respectively. Finally, in Figure~\ref{fig:repair}~(e) we apply the repair network twice and observe that more easy-to-detect low-level artifacts have been introduced.

\subsection{Replication of Real Images}
\label{sec:replica}

Given the corrupt images generated as described in the previous section, we should be ready to train the discriminator $C$. The real images could indeed be just the original images from the same dataset that we used to create the corrupt ones. 
One potential issue with this scheme is that the discriminator may learn to distinguish real from corrupt images based on image processing patterns introduced by the decoder network. 
These patterns may be unnoticeable to the naked eye, but neural networks seem to be quite good at spotting them. For example, networks have learnt to detect chromatic aberration \cite{doersch2015unsupervised} or downsampling patterns \cite{noroozi2017representation}. 

To avoid this issue we use the same autoencoder $\{E,D\}$ used to generate corrupt images, also to replicate real images. Since the same last layer $D_5$ of the decoder is used to generate the real and corrupt images, we expect both images to share the same processing patterns. This should discourage the discriminator from focusing on such patterns.

We therefore pre-train this autoencoder and make sure that it has a high capacity, so that images are replicated with high accuracy. The training of $E$ and $D$ is a standard optimization with the following least-squares objective
\begin{align}
{\cal L}_\text{auto} = \sum_{\mathbf{x}\sim p(\mathbf{x})}|D(E(\mathbf{x})) - \mathbf{x}|^2.
\end{align}

\subsection{Training the Discriminator}
\label{sec:adv}

As just discussed in Sec.~\ref{sec:replica}, to replicate real images we use an autoencoder $\{E,D\}$ and, as described in Sec.~\ref{sec:repair}, to create corrupt images we use a \emph{damage \& repair} autoencoder $\{\Omega \odot E, \hat D\}$, where $\hat D = \{R_1,D_1,R_2,D_2,R_3,D_3,$ $R_4,D_4,R_5,D_5\}$. Then, we train the discriminator and the repair subnetwork $R = \{R_1,R_2,R_3,R_4,R_5\}$ via adversarial training \cite{goodfellow2014generative}.
Our discriminator $C$ has two outputs, a binary probability $C^\text{class}\in [0,1]$ for predicting real vs corrupt images and a prediction mask $C^\text{mask}\in [0,1]^{M\times N}$ to localize artifacts in the image.
Given an image $\mathbf{x}$, training the discriminator $C$ and the repair network $R$  involves solving 
\begin{equation}
\begin{split}
{\cal L}_\text{class} = \min_R \max_C \sum_{\mathbf{x}\sim p(\mathbf{x})}& \log C^\text{class}(D(\phi(\mathbf{x})))\\
 +& \log (1-C^\text{class}(\hat D(\hat \phi(\mathbf{x})))).
\end{split}
\end{equation}
We also train the discriminator to predict the mask $\Omega$ by minimizing
\begin{align}
{\cal L}_\text{mask} = \min_C \sum_{\hat{\mathbf{x}}}\sum_{ij} &\Omega_{ij}\log \sigma\left(C^\text{mask}_{ij}(\hat{\mathbf{x}})\right)\\
& +(1-\Omega_{ij})\log (1-\sigma\left(C^\text{mask}_{ij}(\hat{\mathbf{x}}))\right)\nonumber
\end{align}
where $\hat{\mathbf{x}} = \hat D(\hat \phi(\mathbf{x}))$ and $\sigma(z)= 1/(1+e^{-z})$ is the sigmoid function.


\subsection{Implementation}

Let \texttt{(64)3c2} denote a convolutional layer with 64 filters of size $3\times3$ with a stride of 2. The architecture of the encoder $E$ is then defined by \texttt{(32)3c1}-\texttt{(64)2c2}-\texttt{(128)2c2}-\texttt{(256)2c2}-\texttt{(512)2c2}. The decoder network $D$ is given by \texttt{(256)3rc2}-\texttt{(128)3rc2}-\texttt{(64)3rc2}-\texttt{(32)3rc2}-\texttt{(3)3c1} where $rc$ denotes resize-convolutions (\textit{i.e.}, bilinear resizing followed by standard convolution). 
Batch normalization \cite{ioffe2015batch} is applied at all layers of $E$ and $D$ except the last convolutional layer of $D$. All convolutional layers in $E$ and $D$ are followed by the leaky-ReLU activation $f(x)=\max(x/10, x)$. The filtering of $\phi(x)$ with $u$ is realized using 2D average pooling with a kernel of size $3\times3$.

\begin{figure}[t]
	\begin{center}
  	\includegraphics[width=\linewidth,trim={0cm 0cm 0cm 0cm},clip]{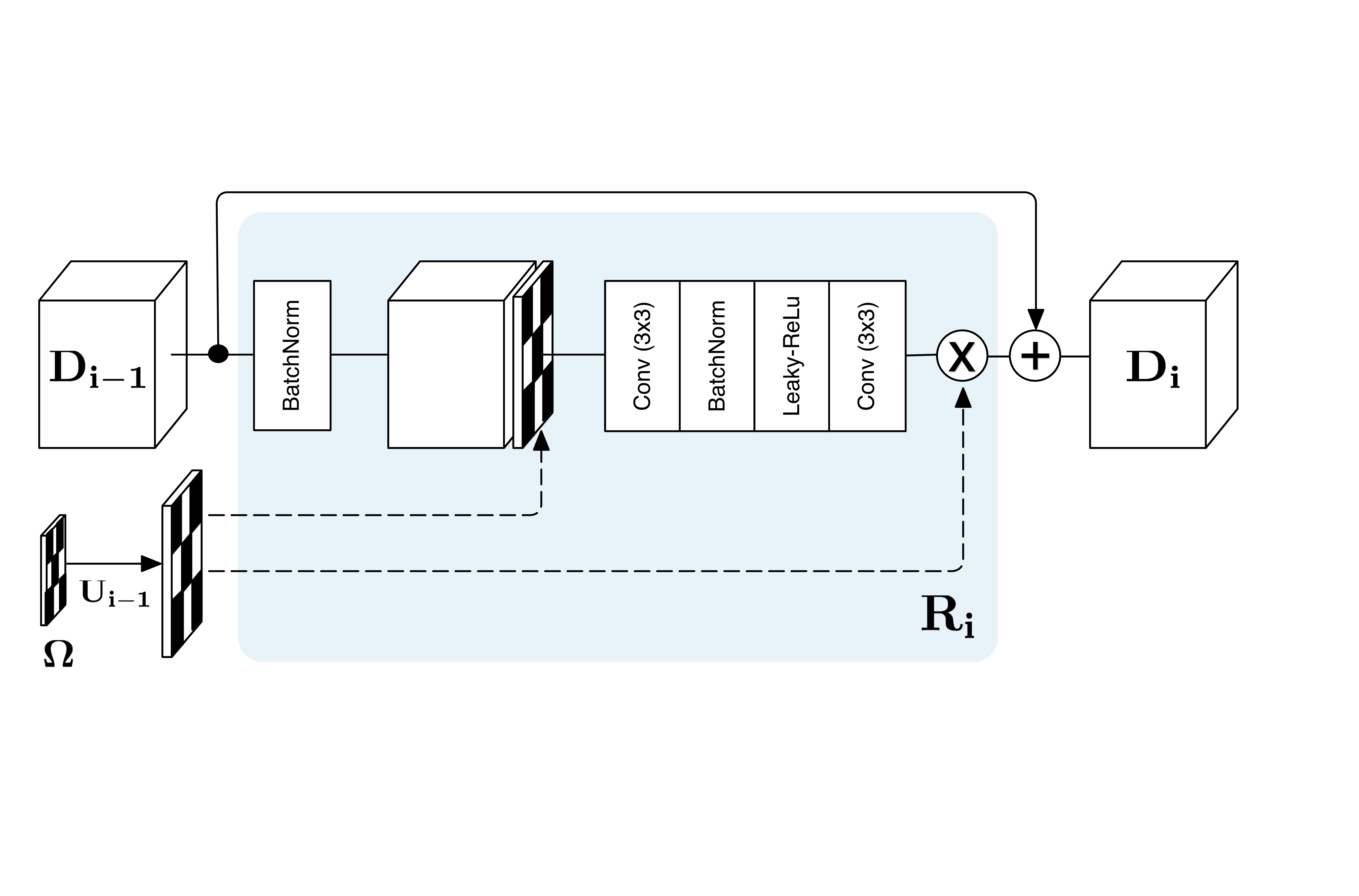}
 	\end{center}
   \caption{The repair block. We use a residual block design. We also concatenate the drop mask to the output of the batch normalization layer as an additional channel before feeding it to the first convolutional layer. We also gate the output of the last convolutional layer with the same drop mask. This ensures that only corrupted regions can be altered by the repair block.}
\label{fig:repairblock}
\end{figure}

The discriminator network $C$ is based on the standard AlexNet architecture \cite{krizhevsky2012imagenet} to allow for a fair comparison with other methods. The network is identical to the original up to \texttt{conv5}. We drop \texttt{pool5} and use a single $3\times3$ convolutional layer for the mask prediction. For the classification we remove the second fully-connected layer. Batch normalization is only applied after \texttt{conv5} during unsupervised training and removed in transfer experiments. The standard ReLU activation is used throughout $C$.
The repair layers follow a similar design as the residual blocks found in ResNet \cite{he2016deep}. Their design is illustrated in Figure~\ref{fig:repairblock}.

Adam \cite{kingma2014adam} with an initial learning rate of $3\cdot10^{-4}$ and the momentum parameter $\beta_1=0.5$ is used for the optimization. We keep all other hyper-parameters at their default values. During training we linearly decay the learning rate to $3\cdot10^{-6}$. The autoencoder is pre-trained for 80 epochs and the \emph{damage \& repair} network is trained for 150 epochs on random $128\times128$ crops of the 1.3M ImageNet \cite{imagenet_cvpr09} training images.

\section{Experiments}


\subsection{Ablation Analysis}

%
%
%
%
%
%
%
%
%
%
%
%
%
%
%
%
%
%

\begin{figure}[t]

\begin{center}\hspace{-3mm}
\resizebox{\linewidth}{!}{
    \begin{subfigure}[]{0.11\linewidth}
        \includegraphics[height=4cm]{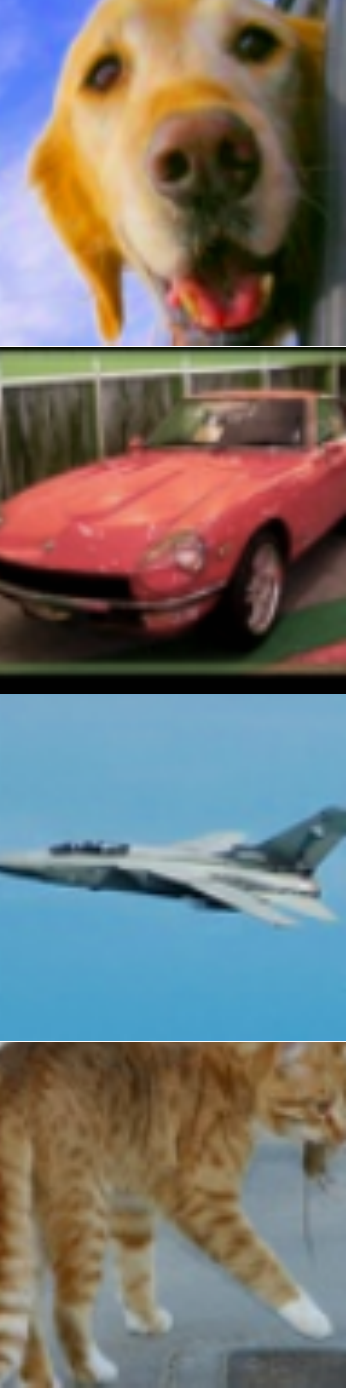}
        \caption*{}\label{fig:gt}
    \end{subfigure}
    \begin{subfigure}[]{0.11\linewidth}
        \includegraphics[height=4cm]{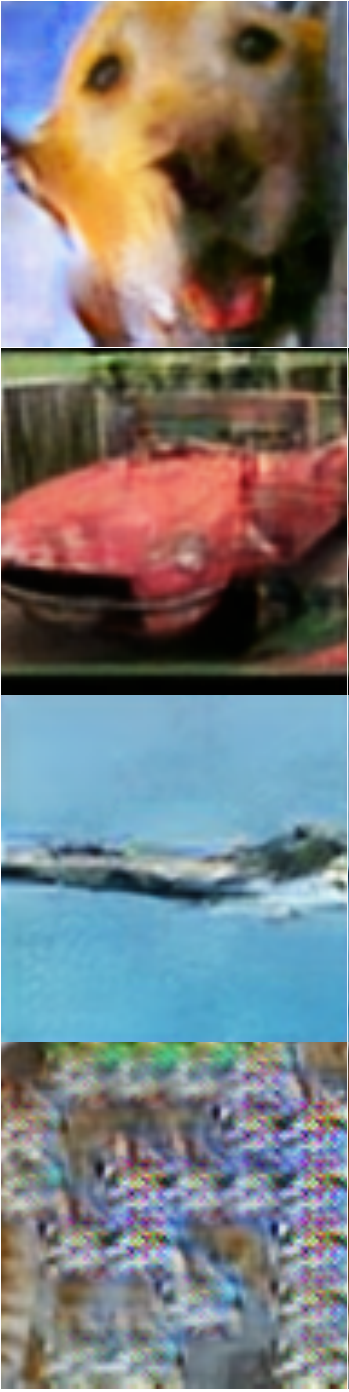}
        \caption{}\label{fig:gt_real}
    \end{subfigure}
	\begin{subfigure}[]{0.11\linewidth}
        \includegraphics[height=4cm]{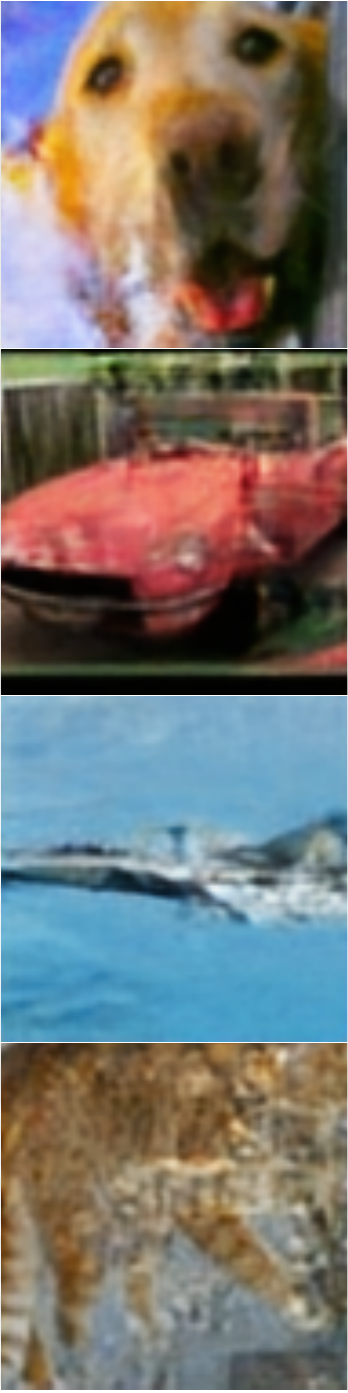}
        \caption{}\label{fig:local}
    \end{subfigure}
    \begin{subfigure}[]{0.11\linewidth}
        \includegraphics[height=4cm]{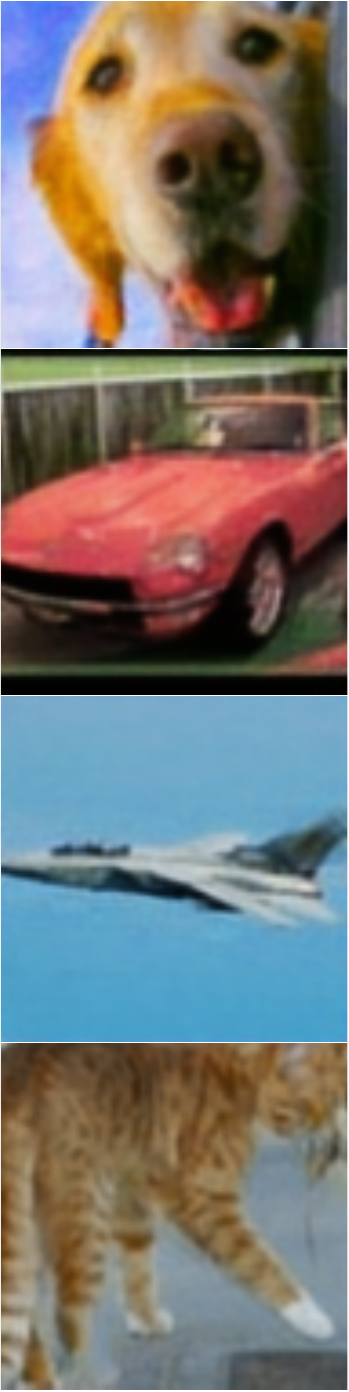}
        \caption{}
    \end{subfigure}    
    \begin{subfigure}[]{0.11\linewidth}
        \includegraphics[height=4cm]{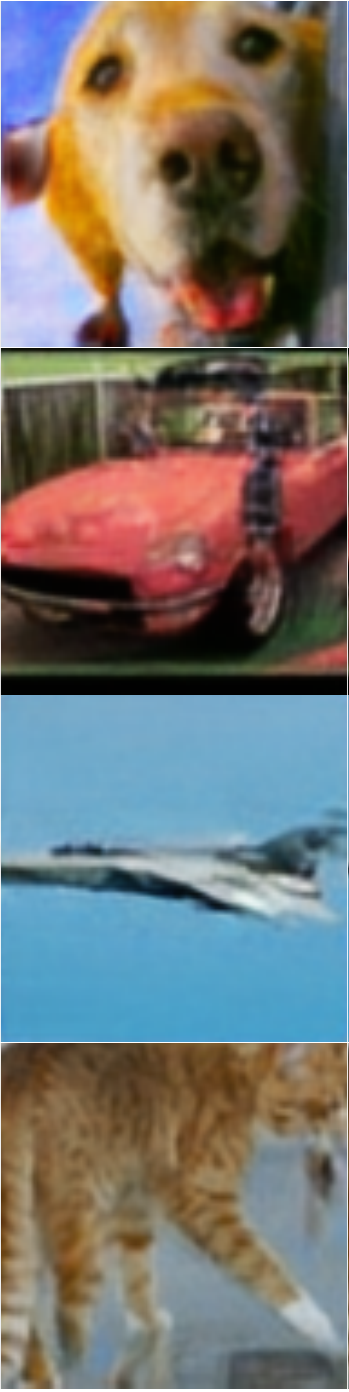}
        \caption{}
    \end{subfigure}
    \begin{subfigure}[]{0.11\linewidth}
        \includegraphics[height=4cm]{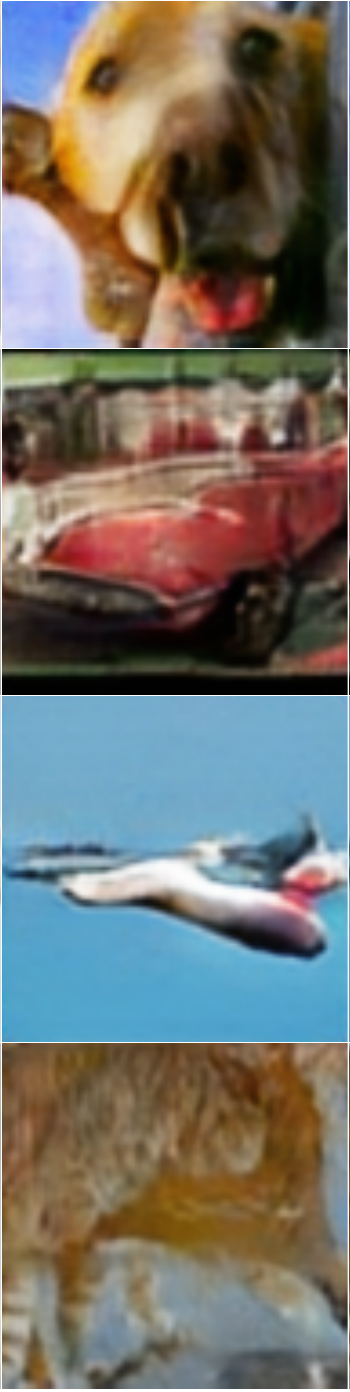}
        \caption{}
    \end{subfigure}
    \begin{subfigure}[]{0.11\linewidth}
        \includegraphics[height=4cm]{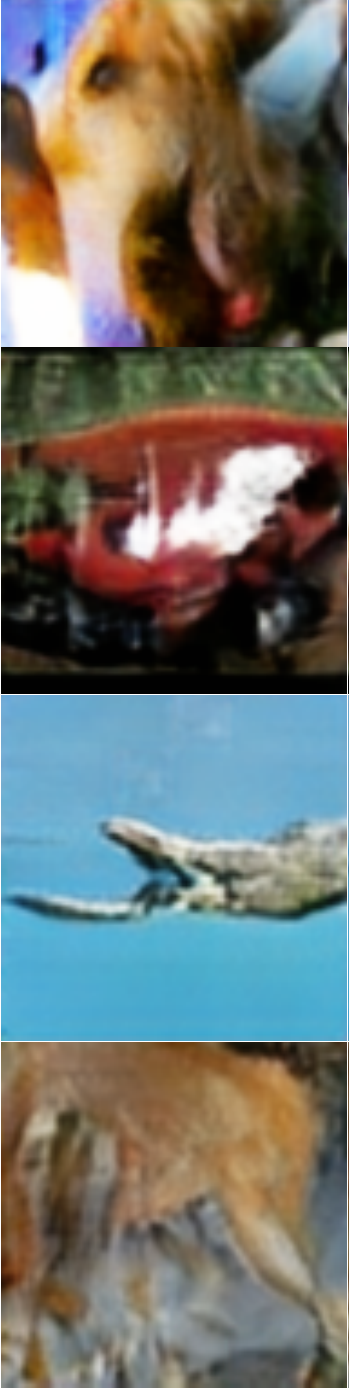}
        \caption{}
    \end{subfigure}}
\end{center}
   \caption{The \emph{damage \& repair} network renderings. The leftmost column shows the input image. Column (a) shows results when input images are used as real examples. Note that this introduces commonly observed GAN artifacts. Column (b) shows results with the local instead of distributed repair network. Columns (c)-(f) show results with dropping rates of 0.1, 0.3, 0.7 and 0.9.}
\label{fig:analysis}
\end{figure}

We validate the main design choices in our model by performing transfer learning for classification on STL-10~\cite{coates2011analysis}. All results were obtained with models trained for 400 epochs on the unlabelled training set and supervised transfer learning for 200 epochs on the labelled training set. During transfer the standard AlexNet architecture is used with the exception that we drop the \texttt{pool5} layer in order to handle the smaller image size. The weights in the convolutional layers are transferred while the fully-connected layers are randomly initialized.
We perform the following set of experiments:
\begin{description}
\setlength\itemsep{-0.1cm}
	\item [(a) Input image as real:] We show that it is better to use autoencoded images as real examples for discriminator training. The rationale here is that the discriminator could exploit the pixel patterns of the decoder network to decide between real and corrupted images. We also observed common GAN artifacts in this setting (see Figure~\ref{fig:analysis}~(a)). 
	In our model, the inputs to the discriminator pass through the same convolutional layer. 
	\item [(b) Distributed vs. local repair network:] Here we illustrate the importance of distributing the repair network throughout the decoder. In the local case we apply the five repair layers consecutively before the first decoder layer. We observe more low-level artifacts in this setup (see Figure~\ref{fig:analysis}~(b)). 
	\item [(c)-(f) Dropping rate:] We show how different values of $\theta$ influence the performance on classification. The following values are considered: $0.1$, $0.3$, $0.5$ (the baseline), $0.7$ and $0.9$. The dropping rate has a strong influence on the performance of the learnt features. Values around $0.7$ give the best results and low dropping rates significantly reduce performance. With low dropping rates it is unlikely that object parts are corrupted. This makes the examples less valuable for learning semantic content.
	\item [(g) Without mask prediction:] This experiment demonstrates that the additional self-supervision signal provided through the drop mask improves performance. 
	\item [(h) $3\times3$ encoder convolutions:] In this experiment we let the encoder features overlap. We replace the $2\times2$ convolutions with $3\times3$ convolutions. This increases the receptive field from $18\times18$ to $33\times33$ and results in an overlap of $15$ pixels. We observe a small decrease in transfer performance. 
	\item [(i) No gating in repair layers:] We demonstrate the influence of training without gating the repair network output with the drop mask. Without gating, the repair network can potentially affect all image regions. 
	\item [(j) No history of corrupted examples:] Following \cite{shrivastava2016learning} we keep a buffer of corrupted examples and build mini-batches where we replace half of the corrupted examples with samples from the buffer. Removing the history has a small negative effect on performance.
	\item [(k) No repair network:] In this ablation we remove the repair network completely. The poor performance in this case illustrates the importance of adversarial training. 
	\item [(l) GAN instead of damage \& repair:] We test the option of training a standard GAN to create the fake examples instead of our damage-repair approach. We observe much poorer performance and unstable adversarial training in this case.  
\end{description}

The resulting transfer learning performance of the discriminator on STL-10 is shown in Table~\ref{tab:stl_arch}. In Figure~\ref{fig:analysis} we show renderings for some of the generative models. 

\begin{table}[t]
\centering
\caption{Influence of different architectural choices on the classification accuracy on STL-10~\cite{coates2011analysis}. Convolutional layers were pre-trained on the proposed self-supervised task and kept fixed during transfer for classification.}\label{tab:stl_arch}
\begin{tabular}{@{}c@{\hspace{0.5em}}l@{\hspace{3.em}}c@{}}
\toprule
& \textbf{Ablation experiment}               			& \textbf{Accuracy} \\ \midrule
\phantom{(bs)} & Baseline (dropping rate = 0.5)			&  79.94\%	\\
(a) & Input image as real	 					&  74.99\%  \\
(b) & Distributed vs. local repair network  		&  77.51\%  \\
(c) & Dropping rate = 0.1				        &  70.92\%	\\
(d) & Dropping rate = 0.3				        &  76.26\%	\\
(e) & Dropping rate = 0.7				        &  81.06\%	\\
(f) & Dropping rate = 0.9				        &  79.60\%	\\
(g) & Without mask prediction	 				&  78.44\%	\\
(h) & $3\times3$ encoder convolutions			&  79.84\%	\\
(i) & No gating in repair layers		   			&  79.66\% 	\\
(j) & No history of corrupted examples   		&  79.76\% 	\\
(k) & No repair network				   			&   54.74\%	\\
(l) & GAN instead of damage \& repair  			&   56.59\%	\\
\bottomrule
\end{tabular}
\end{table}

\begin{table}[t]
\centering
\caption{Comparison of test-set accuracy on STL-10 with other published results. Following the guidelines in \cite{coates2011analysis} the average accuracy from models trained on the ten pre-defined folds is reported. We train a linear classifier on top of \texttt{conv5} features for a fair comparison with the other methods.}
\label{tab:stl}
\resizebox{\linewidth}{!}{%
\begin{tabular}{@{}l@{\hspace{3.em}}c@{\hspace{2.0em}}c@{}}
\toprule
\textbf{Model}                                          			& \textbf{Accuracy} & \textbf{SD}\\ \midrule
Dosovitskiy \etal  \cite{dosovitskiy2014discriminative}      			&  74.2\% & $\pm 0.4$ 	\\
Dundar \etal \etal \cite{dundar2015convolutional} 					&  74.1\% & 	-	\\ 
Huang \etal \cite{Huang_2016_CVPR}  					&  76.8\% & $\pm 0.3$ 	\\ 
Swersky \etal  \cite{swersky2013multi}      			&  70.1\% & $\pm 0.6$ 	\\
Zhao \etal \cite{zhao2015stacked} 					&  74.3\% & 	-	\\ 
Denton \etal \cite{denton2016semi} (finetuned)				&  77.8\% & $\pm 0.8$	\\ \midrule
Ours (\texttt{conv1-conv5} frozen) &  \textbf{76.9\%} &  $\pm 0.2$  \\
Ours (\texttt{conv1-conv5} finetuned) &  \textbf{80.1\%} &  $\pm 0.3$  \\
\bottomrule
\end{tabular}}
\end{table}

\subsection{Transfer Learning Experiments}
We perform transfer learning experiments for classification, detection and semantic segmentation on standard datasets with the AlexNet discriminator pre-trained using a dropping rate of $\theta=0.7$. 

\subsubsection{Classification on STL-10}
The STL-10 dataset \cite{coates2011analysis} is designed with unsupervised representation learning in mind and is a common baseline for comparison. The dataset contains 100,000 \emph{unlabeled} training images, 5,000 \emph{labeled} training images evenly distributed across 10 classes for supervised transfer and 8,000 test images. The data consists of $96\times96$ color images. We randomly resize the images and extract $96\times96$ crops. Unsupervised training is performed for 400 epochs and supervised transfer for an additional 200 epochs.

We follow the standard evaluation protocol and perform supervised training on the ten pre-defined folds. We compare the resulting average accuracy to state-of-the-art results in Table \ref{tab:stl}. We can observe an increase in performance over the other methods. Note that the models compared in Table \ref{tab:stl} do not use the same network architecture, hence making it difficult to attribute the difference in performance to a specific factor. It nonetheless showcases the potential of the proposed self-supervised learning task and model.

\begin{table}[t]
\begin{center}
\caption{Transfer learning results for classification, detection and segmentation on Pascal VOC2007 and VOC2012 compared to state-of-the-art feature learning methods.}\label{tab:voc_layers}
\resizebox{\linewidth}{!}{%
\begin{tabular}{@{}l@{\hspace{.1em}}c@{\hspace{.2em}}c@{\hspace{.5em}}c@{\hspace{.5em}}c@{}}
\toprule
  &   & \textbf{Classification}  & \textbf{Detection}  & \textbf{Segmentation} \\
\textbf{Model} & \textbf{[Ref]}  & \textbf{(mAP)}  & \textbf{(mAP)}  & \textbf{(mIU)} \\ \midrule
Krizhevsky \etal  \cite{krizhevsky2012imagenet} & \cite{zhang2016colorful} & 79.9\% & 56.8\%  &  48.0\%     \\ 
Random & \cite{pathak2016context} & 53.3\% & 43.4\% & 19.8\%     \\  \hline
Agrawal \etal \cite{agrawal2015learning} &	\cite{donahue2016adversarial} & 54.2\% & 43.9\% & -  \\
Bojanowski \etal \cite{bojanowski2017unsupervised} &	 \cite{bojanowski2017unsupervised} & 65.3\% & 49.4\% & -  \\
Doersch \etal \cite{doersch2015unsupervised} & \cite{donahue2016adversarial}	 & 65.3\% & 51.1\% & -     \\
Donahue \etal \cite{donahue2016adversarial} &	\cite{donahue2016adversarial} & 60.1\% & 46.9\% & 35.2\%        \\
Jayaraman \& Grauman \cite{jayaraman2015learning} & \cite{jayaraman2015learning} & - & 41.7\% & -          \\
Kr\"ahenb\"uhl \etal \cite{krahenbuhl2015data} & \cite{krahenbuhl2015data} & 56.6\% & 45.6\% & 32.6\%          \\
Larsson \etal \cite{larsson2017colorproxy} & \cite{larsson2017colorproxy} & 65.9\% & - & \underline{38.0\%}          \\
Noroozi \& Favaro \cite{noroozi2016unsupervised} & \cite{noroozi2016unsupervised} & 67.6\% & \textbf{53.2\%} & 37.6\%   \\
Noroozi \etal \cite{noroozi2017representation} & \cite{noroozi2017representation} & \underline{67.7\%} & 51.4\% & 36.6\%   \\
Owens \etal \cite{owens2016ambient} &	\cite{owens2016ambient} & 61.3\% & 44.0\% & -          \\
Pathak \etal \cite{pathak2016context} &	\cite{pathak2016context}	 & 56.5\% & 44.5\% & 29.7\%          \\
Pathak \etal \cite{pathakCVPR17learning} &	\cite{pathakCVPR17learning}	 & 61.0\% & 52.2\% & -          \\
Wang \& Gupta \cite{wang2015unsupervised} &	 \cite{krahenbuhl2015data}	 & 63.1\% & 47.4\% & -    \\
Zhang \etal \cite{zhang2016colorful} &	\cite{zhang2016colorful}	 & 65.9\% & 46.9\% & 35.6\%       \\
Zhang \etal \cite{zhang2016split} &	\cite{zhang2016split} & 67.1\%  & 46.7\% & 36.0\%       \\ \midrule
Ours   &    -    &   \textbf{69.8\%}    &   \underline{52.5\%}   & \textbf{38.1\%}             \\ \bottomrule
\end{tabular}}
\end{center}
\end{table}

\subsubsection{Multilabel Classification, Detection and Segmentation on PASCAL VOC}
The Pascal VOC2007 and VOC2012 datasets consist of images coming from 20 object classes. It is a relatively challenging dataset due to the high variability in size, pose, and position of objects in the images. These datasets are standard benchmarks for representation learning. 
We only transfer the convolutional layers of our AlexNet based discriminator and randomly initialize the fully-connected layers. The data-dependent rescaling proposed by Kr\"ahenb\"uhl \etal~\cite{krahenbuhl2015data} is used in all experiments as is standard practice. The convolutional layers are fine-tuned, \ie, not frozen. This demonstrates the usefulness of the discriminator weights as an initialization for other tasks. A comparison to the state-of-the-art feature learning methods is shown in Table~\ref{tab:voc_layers}.

\textbf{Classification on VOC2007.}
 For multilabel classification we use the framework provided by Kr\"ahenb\"uhl \etal~\cite{krahenbuhl2015data}. Fine-tuning is performed on random crops of the \emph{'trainval'} dataset. The final predictions are computed as the average prediction of 10 random crops per test image. With a mAP of 69.8\% we achieve a state-of-the-art performance on this task.

\textbf{Detection on VOC2007.}
The Fast-RCNN \cite{girshickICCV15fastrcnn} framework is used for detection. We follow the guidelines in \cite{krahenbuhl2015data} and use multi-scale training and single-scale testing. All other settings are kept at their default values. With a mAP of 52.5\% we achieve the second best result. 

\textbf{Semantic Segmentation on VOC2012.}
We use the standard FCN framework \cite{long2015fully} with default settings. We train for 100,000 iterations using a fixed learning rate of $10^{-4}$. Our discriminator weights achieve a state-of-the-art result with a mean intersection over union (mIU) of 38.1\%.

\begin{table}[t]
\begin{center}
\caption{Validation set accuracy on ImageNet with linear classifiers trained on the frozen convolutional layers after unsupervised pre-training. Results for the other methods are taken from \cite{noroozi2017representation}.}\label{tab:imnet_layers}
\resizebox{\linewidth}{!}{%
\begin{tabular}{@{}l@{\hspace{1em}}c@{\hspace{1em}} c@{\hspace{1em}} c@{\hspace{1em}} c@{\hspace{1em}} c@{}}
\toprule
\textbf{Model\textbackslash Layer} & \texttt{conv1} & \texttt{conv2} &  \texttt{conv3} &  \texttt{conv4} &  \texttt{conv5} \\ \midrule
Krizhevsky \etal  \cite{krizhevsky2012imagenet} & 19.3\% & 36.3\% & 44.2\% & 48.3\%  & 50.5\%  \\
Random  & 11.6\% & 17.1\% & 16.9\% & 16.3\%  & 14.1\%  \\ \midrule
Doersch \etal \cite{doersch2015unsupervised} & 16.2\% & 23.3\% & 30.2\% & 31.7\%  & 29.6\%  \\
Donahue \etal \cite{donahue2016adversarial} & 17.7\% & 24.5\% & 31.0\% & 29.9\%  & 28.0\%  \\
Kr\"ahenb\"uhl \etal \cite{krahenbuhl2015data} & 17.5\% & 23.0\% & 24.5\% & 23.2\%  & 20.6\%  \\  
Noroozi \& Favaro \cite{noroozi2016unsupervised} & \underline{18.2\%} & 28.8\% & 34.0\% & 33.9\%  & 27.1\%  \\
Noroozi \etal \cite{noroozi2017representation} & 18.0\% & \underline{30.6\%} & 34.3\% & 32.5\%  & 25.7\%  \\  
Pathak \etal \cite{pathak2016context} & 14.1\% & 20.7\% & 21.0\% & 19.8\%  & 15.5\%  \\
Zhang \etal \cite{zhang2016colorful} & 13.1\% & 24.8\% & 31.0\% & 32.6\%  & 31.8\%  \\
Zhang \etal \cite{zhang2016split} & 17.7\% & 29.3\% & \underline{35.4\%} & \underline{35.2\%}  & \underline{32.8\%}  \\
\midrule
Ours   &  \textbf{19.5\%} & \textbf{33.3\%}  & \textbf{37.9\%} & \textbf{38.9\%}  & \textbf{34.9\%}  \\ \bottomrule
\end{tabular}}
\end{center}
\end{table}

\begin{table}[t]
\begin{center}
\caption{Validation set accuracy on Places with linear classifiers trained on the frozen convolutional layers after unsupervised pre-training. Results for the other methods are taken from \cite{noroozi2017representation}.}\label{tab:places_layers}
\resizebox{\linewidth}{!}{%
\begin{tabular}{@{}l@{\hspace{1em}}c@{\hspace{1em}} c@{\hspace{1em}} c@{\hspace{1em}} c@{\hspace{1em}} c@{}}
\toprule
\textbf{Model\textbackslash Layer} & \texttt{conv1} & \texttt{conv2} &  \texttt{conv3} &  \texttt{conv4} &  \texttt{conv5} \\ \midrule
Places-labels \etal  \cite{krizhevsky2012imagenet} & 22.1\% & 35.1\% & 40.2\% & 43.3\%  & 44.6\%  \\
ImageNet-labels \etal  \cite{krizhevsky2012imagenet} & 22.7\% & 34.8\% & 38.4\% & 39.4\%  & 38.7\%  \\
Random  & 15.7\% & 20.3\% & 19.8\% & 19.1\%  & 17.5\%  \\ \midrule
Doersch \etal \cite{doersch2015unsupervised} & 19.7\% & 26.7\% & 31.9\% & 32.7\%  & 30.9\%  \\
Donahue \etal \cite{donahue2016adversarial} & 22.0\% & 28.7\% & 31.8\% & 31.3\%  & 29.7\%  \\
Kr\"ahenb\"uhl \etal \cite{krahenbuhl2015data} & 21.4\% & 26.2\% & 27.1\% & 26.1\%  & 24.0\%  \\  
Noroozi \& Favaro \cite{noroozi2016unsupervised} & \underline{23.0\%} & 31.9\% & 35.0\% & 34.2\%  & 29.3\%  \\
Noroozi \etal \cite{noroozi2017representation} & \textbf{23.3\%} & \underline{33.9\%} & \underline{36.3\%} & \underline{34.7\%}  & 29.6\%  \\  
Owens \etal \cite{owens2016ambient} & 19.9\% & 29.3\% & 32.1\% & 28.8\%  & 29.8\%  \\
Pathak \etal \cite{pathak2016context} & 18.2\% & 23.2\% & 23.4\% & 21.9\%  & 18.4\%  \\
Wang \& Gupta \cite{wang2015unsupervised} & 20.1\% & 28.5\% & 29.9\% & 29.7\%  & 27.9\%  \\

Zhang \etal \cite{zhang2016colorful} & 16.0\% & 25.7\% & 29.6\% & 30.3\%  & 29.7\%  \\
Zhang \etal \cite{zhang2016split} & 21.3\% & 30.7\% & 34.0\% & 34.1\%  & \underline{32.5\%}  \\
\midrule
Ours   &  \textbf{23.3\%} & \textbf{34.3\%}  & \textbf{36.9\%} & \textbf{37.3\%}  & \textbf{34.4\%}  \\ \bottomrule
\end{tabular}}
\end{center}
\end{table}

\subsubsection{Layerwise Performance on ImageNet \& Places}

We evaluate the quality of representations learned at different depths of the network with the evaluation framework introduced in \cite{zhang2016colorful}. All convolutional layers are frozen and multinomial logistic regression classifiers are trained on top of them. The outputs of the convolutional layers are resized such that the flattened features are of similar size ($\sim$$9200$). A comparison to other models on ImageNet is given in Table~\ref{tab:imnet_layers}. Our model outperforms all other approaches in this benchmark. Note also that our \texttt{conv1} features perform even slightly better than the supervised counterparts.
To demonstrate that the learnt representations generalize to other input data, the same experiment was also performed on the Places \cite{NIPS2014_5349} dataset. This dataset contains 2.4M images from 205 scene categories.  As can be seen in Table~\ref{tab:places_layers} we outperform all the other methods for layers \texttt{conv2-conv5}. Note also that we achieve the highest overall accuracy with 37.3\%.

\begin{figure}[t]
\begin{center}
     \begin{subfigure}[b]{0.69\linewidth}
        \includegraphics[width=\linewidth]{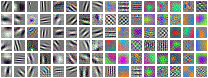}
        \caption{\texttt{conv1} weights}
     \end{subfigure}
     \begin{subfigure}[b]{0.3\linewidth}
        \includegraphics[width=\linewidth]{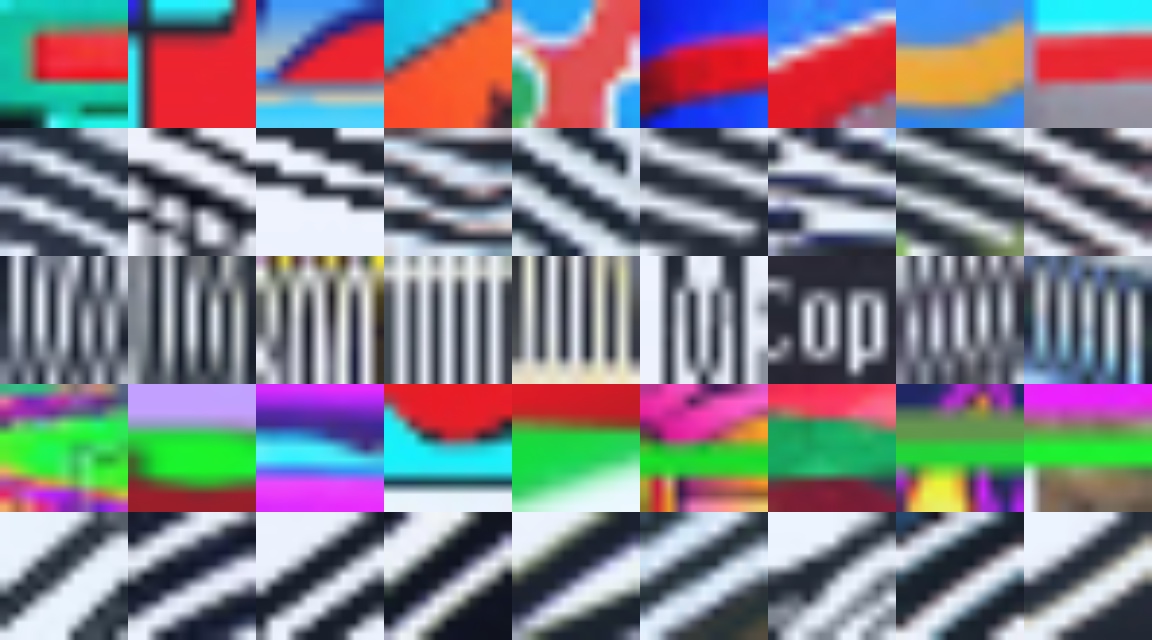}
        \caption{\texttt{conv1}}
     \end{subfigure}
     \begin{subfigure}[b]{0.49\linewidth}
        \includegraphics[width=\linewidth]{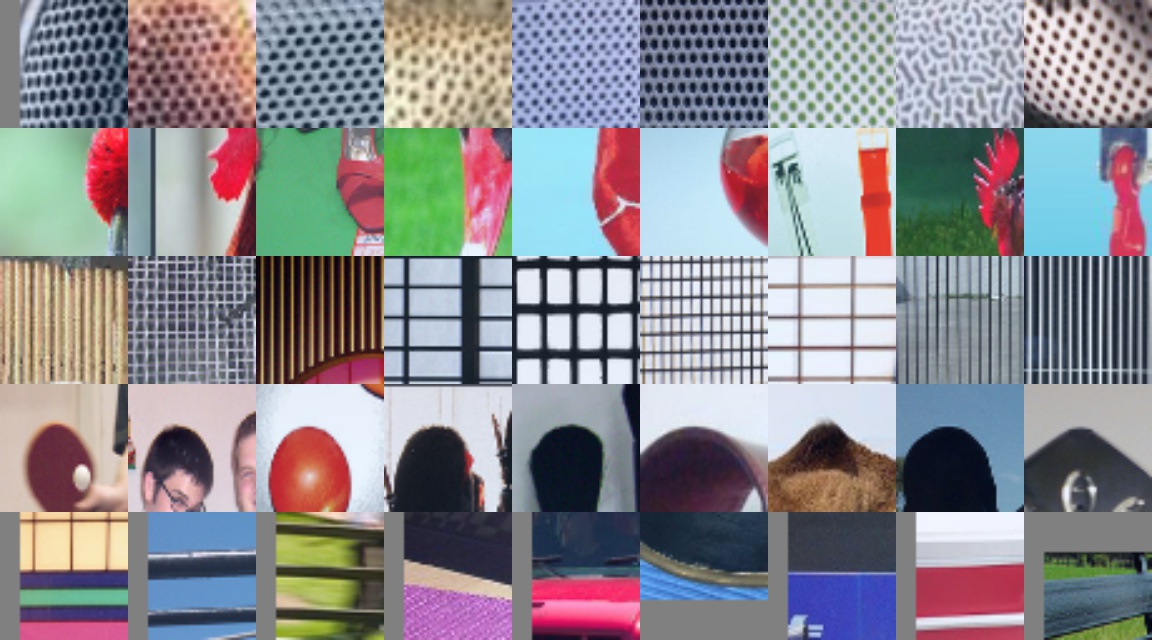}
        \caption{\texttt{conv2}}
    \end{subfigure}
    \begin{subfigure}[b]{0.49\linewidth}
        \includegraphics[width=\linewidth]{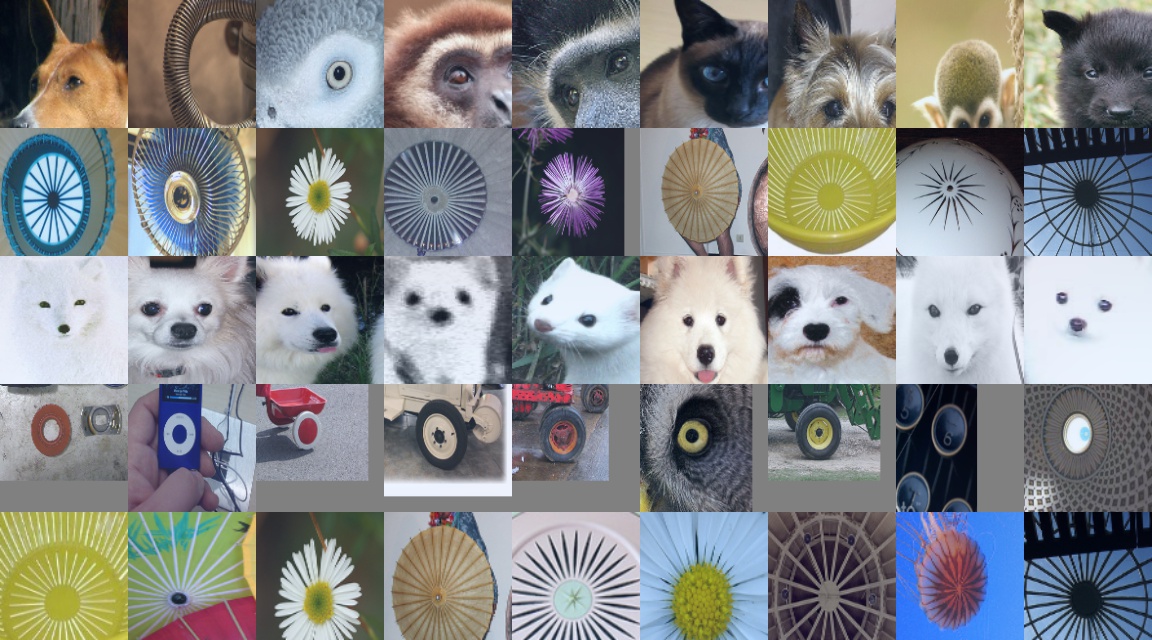}
        \caption{\texttt{conv3}}
    \end{subfigure}
    \begin{subfigure}[b]{0.49\linewidth}
        \includegraphics[width=\linewidth]{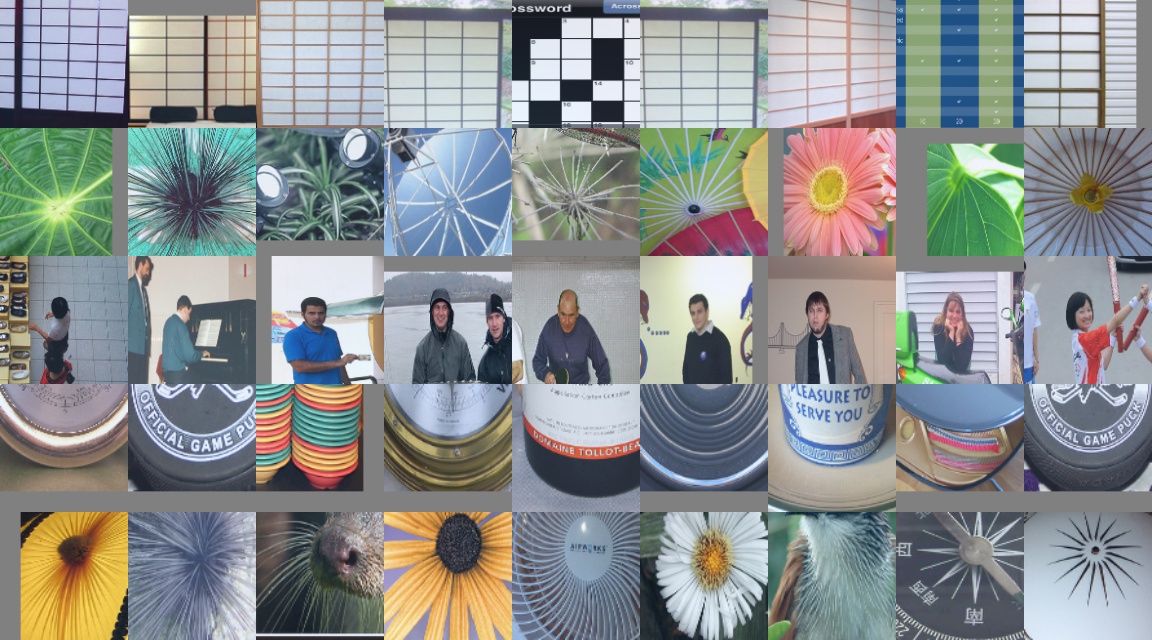}
        \caption{\texttt{conv4}}
    \end{subfigure}
    \begin{subfigure}[b]{0.49\linewidth}
        \includegraphics[width=\linewidth]{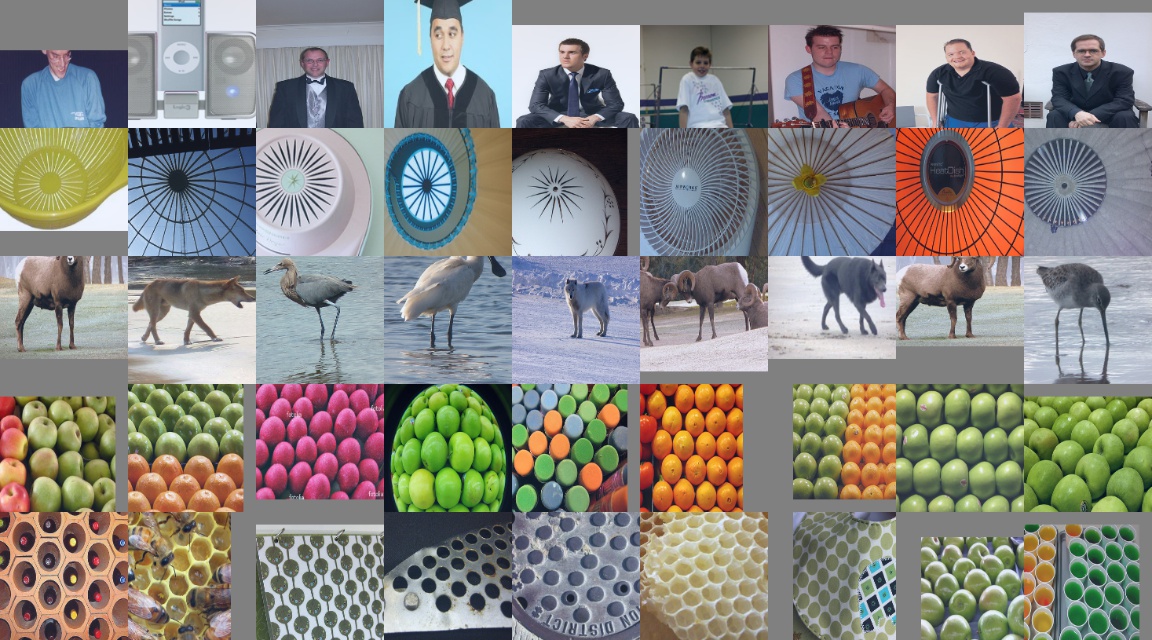}
        \caption{\texttt{conv5}}
    \end{subfigure}
\end{center}
   \caption{Visualisation of the learnt features at different layers of the AlexNet after unsupervised training. We show \texttt{conv1} weights and maximally activating image-patches for five neurons at each layer.}
\label{fig:filter_vis}
\end{figure}

\begin{figure}
\begin{center}
        \includegraphics[width=\linewidth]{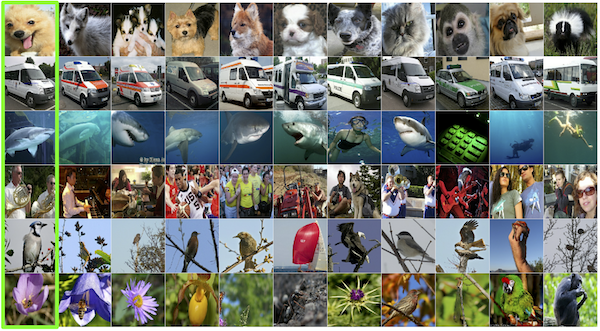}
\end{center}
   \caption{Nearest-neighbor retrievals on the ImageNet validation set obtained with \texttt{conv5} features from the AlexNet discriminator. Nearest-neighbors were computed using cosine-similarity.}
\label{fig:nn_results}
\end{figure}

\begin{figure}[t]
\begin{center}
\setlength{\fboxsep}{0pt}
\setlength{\fboxrule}{1pt}
\fcolorbox{green}{white}{\includegraphics[width=0.19\linewidth,trim={0cm 0cm 0cm 0cm},clip]{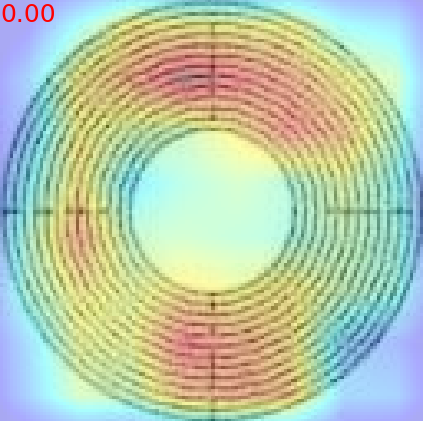}}%
\fcolorbox{green}{white}{\includegraphics[width=0.19\linewidth,trim={0cm 0cm 0cm 0cm},clip]{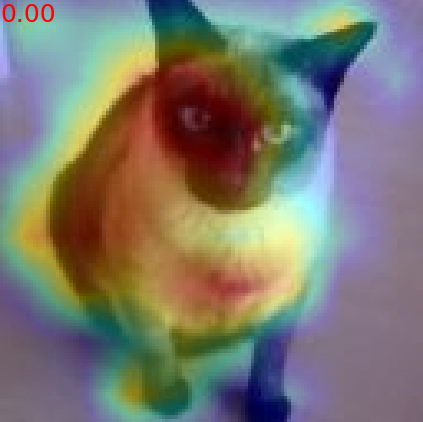}}%
\fcolorbox{green}{white}{\includegraphics[width=0.19\linewidth,trim={0cm 0cm 0cm 0cm},clip]{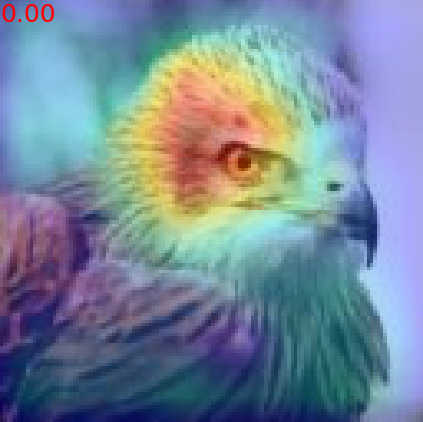}}%
\fcolorbox{green}{white}{\includegraphics[width=0.19\linewidth,trim={0cm 0cm 0cm 0cm},clip]{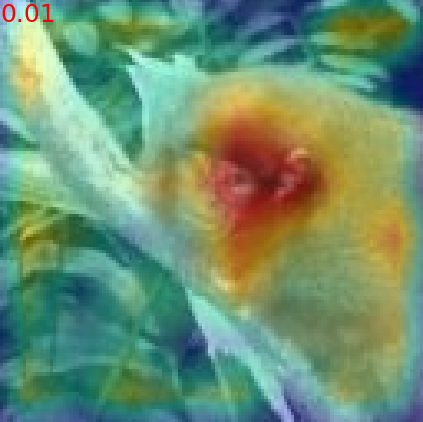}}%
\fcolorbox{green}{white}{\includegraphics[width=0.19\linewidth,trim={0cm 0cm 0cm 0cm},clip]{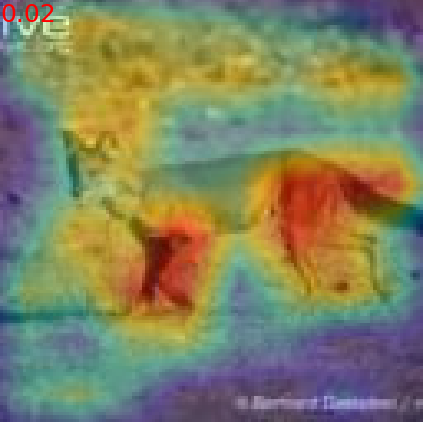}}\\\vspace{-.4mm}%
\fcolorbox{red}{white}{\includegraphics[width=0.19\linewidth,trim={0cm 0cm 0cm 0cm},clip]{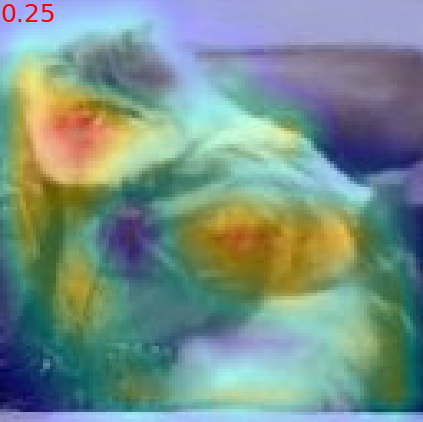}}%
\fcolorbox{red}{white}{\includegraphics[width=0.19\linewidth,trim={0cm 0cm 0cm 0cm},clip]{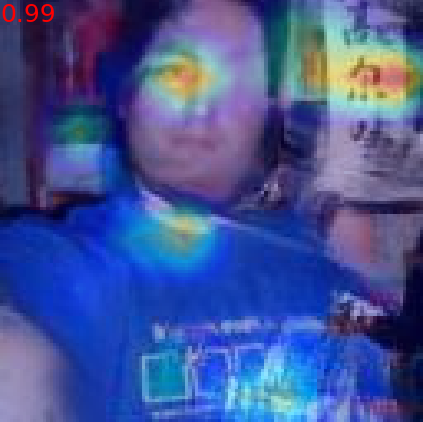}}%
\fcolorbox{red}{white}{\includegraphics[width=0.19\linewidth,trim={0cm 0cm 0cm 0cm},clip]{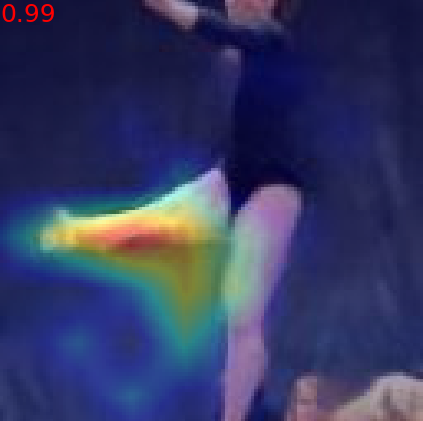}}%
\fcolorbox{red}{white}{\includegraphics[width=0.19\linewidth,trim={0cm 0cm 0cm 0cm},clip]{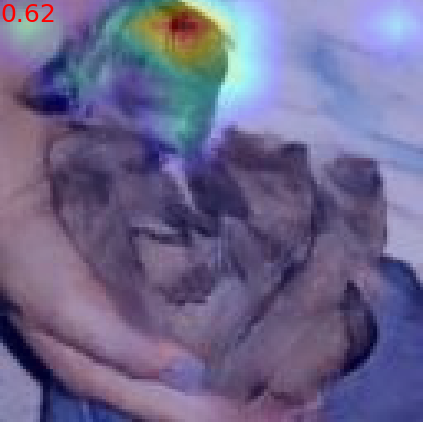}}%
\fcolorbox{red}{white}{\includegraphics[width=0.19\linewidth,trim={0cm 0cm 0cm 0cm},clip]{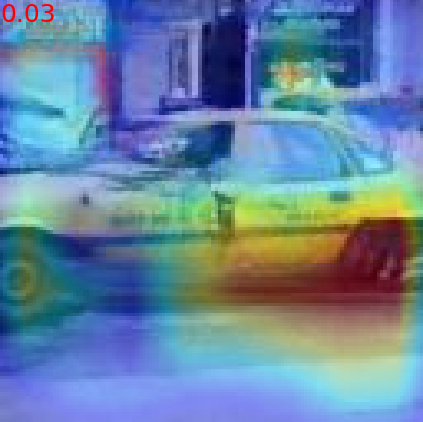}}\\\vspace{-.4mm}%
\end{center}
   \caption{Visualisation of contributing image features for the decision between real images (green border) and images with artifacts (red border). The discriminator appears to focus on object parts such as legs and heads as well as geometric shapes. }
\label{fig:attention}
\end{figure}

\subsection{Qualitative Analysis of the Features}
To better understand what the discriminator has learnt we use different network visualization techniques. We show the learnt \texttt{conv1} filters as well as maximally activating image-patches \cite{girshick14CVPR, yosinski-2015-ICML-DL-understanding-neural-networks} for some neurons of each convolutional layer in Figure~\ref{fig:filter_vis}. We observe prominent edge-detectors in the  \texttt{conv1} filters, much like what can be observed in a supervised AlexNet. Figure \ref{fig:nn_results} shows Nearest-Neighbor retrievals obtained with our \texttt{conv5} features. 

We use Grad-CAM \cite{selvaraju2016grad} in Figure \ref{fig:attention} to illustrate what image regions the discriminator focuses on when making the decision between real and with artifacts. We can observe that the discriminator often looks for missing or existing object parts.

\color{black}

\section{Conclusions}
We have shown how to learn features by classifying images into real or corrupt.
This classification task is designed to use images without human annotation, and thus can exploit large readily-available image datasets.
We have tackled this task with a model that combines autoencoders and an assistive network (the repair network) with adversarial networks. 
The transfer (via fine-tuning) of features learned by the classification network achieves state of the art performance on several benchmarks: ILSVRC2012, Pascal VOC and STL-10.

\textbf{Acknowledgements.} We thank Alexei A. Efros for helpful discussions and feedback. This work was supported by the Swiss National Science Foundation (SNSF) grant number 200021\_169622.

{\small
\bibliographystyle{ieee}
\bibliography{refs}

\begin{thebibliography}{10}\itemsep=-1pt

\bibitem{agrawal2015learning}
P.~Agrawal, J.~Carreira, and J.~Malik.
\newblock Learning to see by moving.
\newblock In {\em Proceedings of the IEEE International Conference on Computer
  Vision}, pages 37--45, 2015.

\bibitem{bertalmio2000image}
M.~Bertalmio, G.~Sapiro, V.~Caselles, and C.~Ballester.
\newblock Image inpainting.
\newblock In {\em Proceedings of the 27th annual conference on Computer
  graphics and interactive techniques}, pages 417--424. ACM
  Press/Addison-Wesley Publishing Co., 2000.

\bibitem{bojanowski2017unsupervised}
P.~Bojanowski and A.~Joulin.
\newblock Unsupervised learning by predicting noise.
\newblock {\em arXiv preprint arXiv:1704.05310}, 2017.

\bibitem{coates2011analysis}
A.~Coates, A.~Ng, and H.~Lee.
\newblock An analysis of single-layer networks in unsupervised feature
  learning.
\newblock In {\em Proceedings of the fourteenth international conference on
  artificial intelligence and statistics}, pages 215--223, 2011.

\bibitem{imagenet_cvpr09}
J.~Deng, W.~Dong, R.~Socher, L.-J. Li, K.~Li, and L.~Fei-Fei.
\newblock Imagenet: A large-scale hierarchical image database.
\newblock In {\em Computer Vision and Pattern Recognition, 2009. CVPR 2009.
  IEEE Conference on}, pages 248--255. IEEE, 2009.

\bibitem{denton2016semi}
E.~Denton, S.~Gross, and R.~Fergus.
\newblock Semi-supervised learning with context-conditional generative
  adversarial networks.
\newblock {\em arXiv preprint arXiv:1611.06430}, 2016.

\bibitem{doersch2015unsupervised}
C.~Doersch, A.~Gupta, and A.~A. Efros.
\newblock Unsupervised visual representation learning by context prediction.
\newblock In {\em Proceedings of the IEEE International Conference on Computer
  Vision}, pages 1422--1430, 2015.

\bibitem{donahue2016adversarial}
J.~Donahue, P.~Kr{\"a}henb{\"u}hl, and T.~Darrell.
\newblock Adversarial feature learning.
\newblock {\em arXiv preprint arXiv:1605.09782}, 2016.

\bibitem{dosovitskiy2014discriminative}
A.~Dosovitskiy, J.~T. Springenberg, M.~Riedmiller, and T.~Brox.
\newblock Discriminative unsupervised feature learning with convolutional
  neural networks.
\newblock In {\em Advances in Neural Information Processing Systems}, pages
  766--774, 2014.

\bibitem{dundar2015convolutional}
A.~Dundar, J.~Jin, and E.~Culurciello.
\newblock Convolutional clustering for unsupervised learning.
\newblock {\em arXiv preprint arXiv:1511.06241}, 2015.

\bibitem{everingham2010pascal}
M.~Everingham, L.~Van~Gool, C.~K. Williams, J.~Winn, and A.~Zisserman.
\newblock The pascal visual object classes (voc) challenge.
\newblock {\em International journal of computer vision}, 88(2):303--338, 2010.

\bibitem{girshickICCV15fastrcnn}
R.~Girshick.
\newblock Fast r-cnn.
\newblock In {\em Proceedings of the IEEE International Conference on Computer
  Vision}, pages 1440--1448, 2015.

\bibitem{girshick14CVPR}
R.~Girshick, J.~Donahue, T.~Darrell, and J.~Malik.
\newblock Rich feature hierarchies for accurate object detection and semantic
  segmentation.
\newblock In {\em Proceedings of the IEEE conference on computer vision and
  pattern recognition}, pages 580--587, 2014.

\bibitem{goodfellow2014generative}
I.~Goodfellow, J.~Pouget-Abadie, M.~Mirza, B.~Xu, D.~Warde-Farley, S.~Ozair,
  A.~Courville, and Y.~Bengio.
\newblock Generative adversarial nets.
\newblock In {\em Advances in Neural Information Processing Systems}, pages
  2672--2680, 2014.

\bibitem{he2016deep}
K.~He, X.~Zhang, S.~Ren, and J.~Sun.
\newblock Deep residual learning for image recognition.
\newblock In {\em Proceedings of the IEEE Conference on Computer Vision and
  Pattern Recognition}, pages 770--778, 2016.

\bibitem{hinton1994autoencoders}
G.~E. Hinton and R.~S. Zemel.
\newblock Autoencoders, minimum description length, and helmholtz free energy.
\newblock {\em Advances in neural information processing systems}, pages 3--3,
  1994.

\bibitem{Huang_2016_CVPR}
C.~Huang, C.~Change~Loy, and X.~Tang.
\newblock Unsupervised learning of discriminative attributes and visual
  representations.
\newblock In {\em Proceedings of the IEEE Conference on Computer Vision and
  Pattern Recognition}, pages 5175--5184, 2016.

\bibitem{ioffe2015batch}
S.~Ioffe and C.~Szegedy.
\newblock Batch normalization: Accelerating deep network training by reducing
  internal covariate shift.
\newblock In {\em International Conference on Machine Learning}, pages
  448--456, 2015.

\bibitem{jayaraman2015learning}
D.~Jayaraman and K.~Grauman.
\newblock Learning image representations tied to ego-motion.
\newblock In {\em Proceedings of the IEEE International Conference on Computer
  Vision}, pages 1413--1421, 2015.

\bibitem{kingma2014adam}
D.~Kingma and J.~Ba.
\newblock Adam: A method for stochastic optimization.
\newblock {\em arXiv preprint arXiv:1412.6980}, 2014.

\bibitem{kingma2013auto}
D.~P. Kingma and M.~Welling.
\newblock Auto-encoding variational bayes.
\newblock {\em arXiv preprint arXiv:1312.6114}, 2013.

\bibitem{krahenbuhl2015data}
P.~Kr{\"a}henb{\"u}hl, C.~Doersch, J.~Donahue, and T.~Darrell.
\newblock Data-dependent initializations of convolutional neural networks.
\newblock {\em arXiv preprint arXiv:1511.06856}, 2015.

\bibitem{krizhevsky2012imagenet}
A.~Krizhevsky, I.~Sutskever, and G.~E. Hinton.
\newblock Imagenet classification with deep convolutional neural networks.
\newblock In {\em Advances in neural information processing systems}, pages
  1097--1105, 2012.

\bibitem{larsson2017colorproxy}
G.~Larsson, M.~Maire, and G.~Shakhnarovich.
\newblock Colorization as a proxy task for visual understanding.
\newblock In {\em CVPR}, 2017.

\bibitem{long2015fully}
J.~Long, E.~Shelhamer, and T.~Darrell.
\newblock Fully convolutional networks for semantic segmentation.
\newblock In {\em Proceedings of the IEEE Conference on Computer Vision and
  Pattern Recognition}, pages 3431--3440, 2015.

\bibitem{makhzani2015adversarial}
A.~Makhzani, J.~Shlens, N.~Jaitly, and I.~Goodfellow.
\newblock Adversarial autoencoders.
\newblock {\em arXiv preprint arXiv:1511.05644}, 2015.

\bibitem{misra2016shuffle}
I.~Misra, C.~L. Zitnick, and M.~Hebert.
\newblock Shuffle and learn: unsupervised learning using temporal order
  verification.
\newblock In {\em European Conference on Computer Vision}, pages 527--544.
  Springer, 2016.

\bibitem{noroozi2016unsupervised}
M.~Noroozi and P.~Favaro.
\newblock Unsupervised learning of visual representations by solving jigsaw
  puzzles.
\newblock In {\em European Conference on Computer Vision}, pages 69--84.
  Springer, 2016.

\bibitem{noroozi2017representation}
M.~Noroozi, H.~Pirsiavash, and P.~Favaro.
\newblock Representation learning by learning to count.
\newblock {\em arXiv preprint arXiv:1708.06734}, 2017.

\bibitem{owens2016ambient}
A.~Owens, J.~Wu, J.~H. McDermott, W.~T. Freeman, and A.~Torralba.
\newblock Ambient sound provides supervision for visual learning.
\newblock In {\em European Conference on Computer Vision}, pages 801--816.
  Springer, 2016.

\bibitem{pathakCVPR17learning}
D.~Pathak, R.~Girshick, P.~Doll\'{a}r, T.~Darrell, and B.~Hariharan.
\newblock Learning features by watching objects move.
\newblock In {\em Computer Vision and Pattern Recognition ({CVPR})}, 2017.

\bibitem{pathak2016context}
D.~Pathak, P.~Krahenbuhl, J.~Donahue, T.~Darrell, and A.~A. Efros.
\newblock Context encoders: Feature learning by inpainting.
\newblock In {\em Proceedings of the IEEE Conference on Computer Vision and
  Pattern Recognition}, pages 2536--2544, 2016.

\bibitem{pinto2016curious}
L.~Pinto, D.~Gandhi, Y.~Han, Y.-L. Park, and A.~Gupta.
\newblock The curious robot: Learning visual representations via physical
  interactions.
\newblock In {\em European Conference on Computer Vision}, pages 3--18.
  Springer, 2016.

\bibitem{radford2015unsupervised}
A.~Radford, L.~Metz, and S.~Chintala.
\newblock Unsupervised representation learning with deep convolutional
  generative adversarial networks.
\newblock {\em arXiv preprint arXiv:1511.06434}, 2015.

\bibitem{renNIPS15fasterrcnn}
S.~Ren, K.~He, R.~Girshick, and J.~Sun.
\newblock Faster r-cnn: Towards real-time object detection with region proposal
  networks.
\newblock In {\em Advances in neural information processing systems}, pages
  91--99, 2015.

\bibitem{salimans2016improved}
T.~Salimans, I.~Goodfellow, W.~Zaremba, V.~Cheung, A.~Radford, and X.~Chen.
\newblock Improved techniques for training gans.
\newblock In {\em Advances in Neural Information Processing Systems}, pages
  2226--2234, 2016.

\bibitem{selvaraju2016grad}
R.~R. Selvaraju, A.~Das, R.~Vedantam, M.~Cogswell, D.~Parikh, and D.~Batra.
\newblock Grad-cam: Why did you say that? visual explanations from deep
  networks via gradient-based localization.
\newblock {\em arXiv preprint arXiv:1610.02391}, 2016.

\bibitem{sharif2014cnn}
A.~Sharif~Razavian, H.~Azizpour, J.~Sullivan, and S.~Carlsson.
\newblock Cnn features off-the-shelf: an astounding baseline for recognition.
\newblock In {\em Proceedings of the IEEE Conference on Computer Vision and
  Pattern Recognition Workshops}, pages 806--813, 2014.

\bibitem{shrivastava2016learning}
A.~Shrivastava, T.~Pfister, O.~Tuzel, J.~Susskind, W.~Wang, and R.~Webb.
\newblock Learning from simulated and unsupervised images through adversarial
  training.
\newblock {\em arXiv preprint arXiv:1612.07828}, 2016.

\bibitem{swersky2013multi}
K.~Swersky, J.~Snoek, and R.~P. Adams.
\newblock Multi-task bayesian optimization.
\newblock In {\em Advances in neural information processing systems}, pages
  2004--2012, 2013.

\bibitem{vincent2010stacked}
P.~Vincent, H.~Larochelle, I.~Lajoie, Y.~Bengio, and P.-A. Manzagol.
\newblock Stacked denoising autoencoders: Learning useful representations in a
  deep network with a local denoising criterion.
\newblock {\em Journal of Machine Learning Research}, 11(Dec):3371--3408, 2010.

\bibitem{wang2015unsupervised}
X.~Wang and A.~Gupta.
\newblock Unsupervised learning of visual representations using videos.
\newblock In {\em Proceedings of the IEEE International Conference on Computer
  Vision}, pages 2794--2802, 2015.

\bibitem{yosinski-2015-ICML-DL-understanding-neural-networks}
J.~Yosinski, J.~Clune, A.~Nguyen, T.~Fuchs, and H.~Lipson.
\newblock Understanding neural networks through deep visualization.
\newblock In {\em Deep Learning Workshop, International Conference on Machine
  Learning (ICML)}, 2015.

\bibitem{zhang2016colorful}
R.~Zhang, P.~Isola, and A.~A. Efros.
\newblock Colorful image colorization.
\newblock In {\em European Conference on Computer Vision}, pages 649--666.
  Springer, 2016.

\bibitem{zhang2016split}
R.~Zhang, P.~Isola, and A.~A. Efros.
\newblock Split-brain autoencoders: Unsupervised learning by cross-channel
  prediction.
\newblock In {\em CVPR}, 2017.

\bibitem{zhao2015stacked}
J.~Zhao, M.~Mathieu, R.~Goroshin, and Y.~Lecun.
\newblock Stacked what-where auto-encoders.
\newblock {\em arXiv preprint arXiv:1506.02351}, 2015.

\bibitem{NIPS2014_5349}
B.~Zhou, A.~Lapedriza, J.~Xiao, A.~Torralba, and A.~Oliva.
\newblock Learning deep features for scene recognition using places database.
\newblock In Z.~Ghahramani, M.~Welling, C.~Cortes, N.~D. Lawrence, and K.~Q.
  Weinberger, editors, {\em Advances in Neural Information Processing Systems
  27}, pages 487--495. Curran Associates, Inc., 2014.

\end{thebibliography}
}
\end{document}